\documentclass{article}

\usepackage{microtype}
\usepackage{graphicx}
\usepackage{subcaption}
\usepackage{booktabs} 
\usepackage{amsmath}
\usepackage{amsfonts}
\usepackage{dsfont}
\usepackage{xcolor}
\usepackage{xspace}

\usepackage{hyperref}

\DeclareMathOperator*{\argminA}{arg\,min}

\newcommand\bertbase{BERT$_{\small \textsc{BASE}}$\xspace}
\newcommand\bertlarge{BERT$_{\small \textsc{LARGE}}$\xspace}
\newcommand\robertabase{RoBERTa$_{\small \textsc{BASE}}$\xspace}

\usepackage[accepted]{icml2023}

\usepackage{amsmath}
\usepackage{amssymb}
\usepackage{mathtools}
\usepackage{amsthm}

\usepackage[capitalize,noabbrev]{cleveref}

\theoremstyle{plain}

\theoremstyle{definition}

\theoremstyle{remark}

\usepackage[textsize=tiny]{todonotes}

\icmltitlerunning{Backward Compatibility During Data Updates by Weight Interpolation}

\begin{document}

\twocolumn[
\icmltitle{Backward Compatibility During Data Updates by Weight Interpolation}




\begin{icmlauthorlist}
\icmlauthor{Raphael Schumann\textsuperscript{*}}{hd}
\icmlauthor{Elman Mansimov}{aws}
\icmlauthor{Yi-An Lai}{aws}
\icmlauthor{Nikolaos Pappas}{aws}
\icmlauthor{Xibin Gao}{aws}
\icmlauthor{Yi Zhang}{aws}
\end{icmlauthorlist}

\icmlaffiliation{hd}{Computational Linguistics, Heidelberg University, Germany}
\icmlaffiliation{aws}{AWS AI Labs}

\icmlcorrespondingauthor{Elman Mansimov}{mansimov@amazon.com}

\icmlkeywords{Machine Learning, ICML}

\vskip 0.3in
]



\printAffiliationsAndNotice{\textsuperscript{*}Work done while interning at AWS AI Labs.} 

\begin{abstract}
Backward compatibility of model predictions is a desired property when updating a machine learning driven application. It allows to seamlessly improve the underlying model without introducing regression bugs. In classification tasks these bugs occur in the form of negative flips. This means an instance that was correctly classified by the old model is now classified incorrectly by the updated model. This has direct negative impact on the user experience of such systems e.g. a frequently used voice assistant query is suddenly misclassified.
A common reason to update the model is when new training data becomes available and needs to be incorporated. Simply retraining the model with the updated data introduces the unwanted negative flips. We study the problem of regression during data updates and propose Backward Compatible Weight Interpolation~(BCWI). This method interpolates between the weights of the old and new model and we show in extensive experiments that it reduces negative flips without sacrificing the improved accuracy of the new model. BCWI is straight forward to implement and does not increase inference cost. We also explore the use of importance weighting during interpolation and averaging the weights of multiple new models in order to further reduce negative flips.
\end{abstract}

\section{Introduction}
\begin{figure}[ht]
    \centering
    \includegraphics[width=0.39\textwidth]{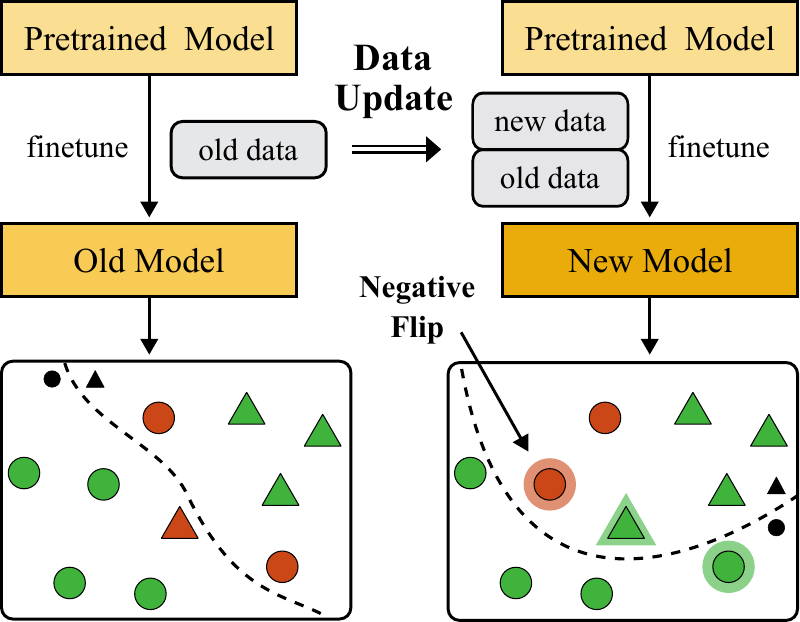}
    \caption{The left column shows the common workflow of finetuning a pretrained model on a given dataset in order to learn a classifier. A data update occurs when new data becomes available and is added to the existing data. The right column depicts the finetuning of the pretrained model on the updated (old and new) data. After expanding the training set, the new model makes higher number of correct predictions compared to the old model. Despite this, the prediction of some instances are flipped from the correct label to an incorrect one. These so called regression errors hinder the adoption of the new model. Our work proposes a method to reduce those negative flips during data updates.}
    \label{fig:motivation}
\end{figure}
In conventional software development it is established routine to identify and fix regression bugs before deploying a new version. Regression bugs describe defects in already existing features and are particularly sensible for end users because accustomed workflows are affected. In machine learning driven applications however the main focus usually lies on improving the underlying model and regression is rarely measured, let alone actively mitigated. This prevents backward compatibility of e.g. visual search systems~\cite{Shen2020TowardsBR} or virtual voice assistants~\cite{Cai2022MeasuringAR}. Previous work on mitigating regression in machine learning models focuses on cases where the model architecture~\cite{Yan2021PositiveCongruentTT, Cai2022MeasuringAR} or pretraining procedure~\cite{Xie2021RegressionBA} is updated. For example, updating a finetuned BERT model~\cite{devlin-etal-2019-bert} to a RoBERTa based model~\cite{Liu2019RoBERTaAR} which is finetuned on the same task specific data. Such fundamental modifications are done rather infrequently and it is more common to update the training data of a model in order to improve a deployed system. One such type of machine learning based system that undergoes frequent data updates are virtual assistants and chatbots. A data update consists of additional labeled utterances and commonly aims to improve classification performance or to support new classes. Training a new model on the updated data introduces regression in the form of negative flips. As depicted in Figure~\ref{fig:motivation}, a negative flip is a data point that was correctly classified by the old model and is now classified incorrectly by the new model. This happens despite the overall better accuracy of the new model. From a user's perspective it seems as if the virtual assistant or chatbot got worse because familiar utterances are suddenly misinterpreted. On the other hand, the overall better accuracy is only perceived over time. The negative user impact of regression and abundance of data updates motivates us to study the mitigation of regression during data updates in multi-class text classification. The outlined data update setting can be categorized as a continual learning problem. But in two key aspects it is distinct from the commonly studied settings~\cite{PARISI201954}. (i) We assume full access to the old training data. As such, catastrophic forgetting in terms of accuracy drop is avoided by joint training. (ii) We instead measure forgetting/interference by number of negative flips between old model and new model.

To reduce negative flips during data updates, we propose Backward Compatible Weight Interpolation~(BCWI) in this paper. BCWI describes the interpolation between the weights of the old model and the weights of the new model. The interpolation largely recovers the prediction pattern of the old model without hurting the improved accuracy of the new model. The method is informed by recent success of weight interpolation for robust finetuning~\cite{Wortsman2022RobustFO} and model patching~\cite{Ilharco2022PatchingOM}.
While these works focus on avoiding catastrophic forgetting in terms of task accuracy, we are interested in reducing negative flips while maintaining high accuracy.
We further introduce FisherBCWI which uses the Fisher information matrix as importance weighting \cite{Kirkpatrick2017OvercomingCF, Matena2021MergingMW} and SoupBCWI which employs soup ensembles~\cite{Wortsman2022ModelSA} to further reduce negative flips. The proposed methods do not modify the training process and do not increase inference cost. We describe BCWI and its variants in detail and empirically show on three datasets and two update scenarios (adding i.i.d. data and adding new classes) that they reduce negative flips by up to three times while maintaining the improved accuracy of the new model. This property of weight interpolation has not been explored before and constitutes a substantial step towards regression free data updates.

\section{Related Work}
\subsection{Mitigating Regression}
Previous work focuses on reducing negative flips when updating the model architecture or the pretraining procedure. In these settings the available data is static and not affected by the update as in our work.
\citet{Yan2021PositiveCongruentTT} propose focal distillation which trains the new model to make similar predictions as the old model by minimizing KL divergence of model predictions. Compared to regular knowledge distillation~\cite{Hinton2015DistillingTK} the focal distillation method applies a higher weight to training examples that are correctly predicted by the old model. This method produces less negative flips when updating model architectures in image classification tasks.~\citet{Xie2021RegressionBA} adopt this technique and show its applicability to text based models, e.g. updating \bertbase to \bertlarge or to RoBERTa. \citet{Yan2021PositiveCongruentTT} and \citet{Xie2021RegressionBA} report that ensembling multiple new models reduces negative flips. The latter propose to avoid the higher inference cost by selecting a single centric model that best represents the predictions of the ensemble. \citet{Zhao2022ELODIEL} propose ensemble logit difference inhibition (ELODI), a technique to distill the ensemble of new models into a single model. We instead employ the more straight forward soup ensemble~\cite{Wortsman2022ModelSA} that produces a single model by averaging the weights of all models in the ensemble. We find that the soup ensemble of new models is as good as a probability ensemble in reducing negative flips, but without increasing the inference cost.
\citet{Cai2022MeasuringAR} tackles regression in structured prediction tasks by using the old model to rerank outputs of the new model.

\subsection{Weight Interpolation}
Weight interpolation and weight averaging are known to improve classification performance in different settings. Averaging the weights of multiple model checkpoints along a cyclic learning rate schedule leads to better classification generalization~\cite{Izmailov2018AveragingWL}. Averaging the weights of multiple models, initialized by the same pretrained model and finetuned with different hyperparameter, improves accuracy in classification tasks~\cite{Wortsman2022ModelSA} and out-of-distribution generalization~\cite{rame2022diwa}. Weight interpolation is also used to merge the task specific accuracy of a finetuned model with the zero-shot capability of its ancestor model~\cite{Wortsman2022RobustFO, Ilharco2022PatchingOM}. Looking beyond simple averaging, \citet{Matena2021MergingMW} use the Fisher information matrix to scale each model weight by importance. We use the same importance weighting for the FisherBCWI method. To the best of our knowledge, we are the first to explore weight interpolation for mitigating data update regression.

\subsection{Continual Learning}
Continual learning studies the problem of incrementally adding new knowledge to a model while avoiding catastrophic forgetting~\cite{Ratcliff1990-RATCMO, MCCLOSKEY1989109}.
Knowledge arrives in the form of new tasks, additional classes or data with shifted distribution~\cite{DeLange2022ACL}. Catastrophic forgetting is measured by accuracy drop on previous data and tasks. It arises from the imposed constraint that one has none or limited access to previous data when new knowledge is incorporated. The constraint is motivated by analogy of how humans learn over time~\cite{MCCLOSKEY1989109} or storage feasibility~\cite{Sodhani2022AnIT}. We instead allow access to the old data because the amount is manageable and we do not focus on simulating lifelong learning. This setting reinforces the need to measure catastrophic forgetting and interference not only in terms of overall accuracy, but also in terms of reducing negative flip rate.

One set of methods used to avoid catastrophic forgetting in continual learning aims at memorizing only a subset of the previous information~\cite{LopezPaz2017GradientEM, Guo2020ImprovedSF, Rebuffi2017iCaRLIC}. Others tackle the problem by weight regularization, preventing the model weights to deviate too far from the old model. Prior Weight Decay \cite{Wiese2017NeuralDA, Lee2020MixoutER} moves the current model weights in the direction of the weights of the old model during each training step. Mixout~\cite{Lee2020MixoutER} randomly replaces a subset of the current weights with the weights of the old model at each training step. \citet{Kirkpatrick2017OvercomingCF} introduce Elastic Weight Consolidation~(EWC) which uses the diagonal Fisher information matrix to weigh the importance of each model parameter in L2 regularization. We show in our experiments that the weight regularization techniques also reduce the number of negative flips.
In model editing~\cite{Mitchell2021FastME, de-cao-etal-2021-editing}, a meta-model is learned to directly modify the learned weights in order to individually correct outdated or factually incorrect instances.

\section{Problem Formulation: Regression in Data Updates}
\label{sec:problem}
In order to measure regression in classification models, \citet{Yan2021PositiveCongruentTT} introduced negative flip rate:
\begin{equation}
\mathrm{NFR} = \frac{1}{N} \sum_{i}^{N} \mathds{1}[ f_{\theta_{old}}(x_i)=y_i \land f_{\theta_{new}}(x_i) \neq y_i],
\label{equ:nfr}
\end{equation}
where $f_{\theta_{old}}$ is the old model and $f_{\theta_{new}}$ the new, updated model. NFR is measured on a given regression set with $N$ input ($x$) and label ($y$) pairs.  Negative flips are instances that are predicted correctly by the old model and are incorrectly predicted by the new model. Consequently NFR is the ratio of negative flips to the total number of instances in the regression set i.e. the development or test set.

We formulate the problem of minimizing regression during data updates in the following way. A deployed model with weights $\theta_{old}$ was trained by finetuning a pretrained model $\theta_{pre}$ on currently available data~$D_{old}$:
\begin{equation}
\theta_{old} = \argminA_{\theta} \mathcal{L}(\theta | \theta_{pre}, \mathcal{D}_{old}),
\label{equ:old}
\end{equation}
where $\mathcal{L}$ is the classification loss. We now obtain additional data and update the available data to get \mbox{$\mathcal{D}_{upd} = \mathcal{D}_{old} \cup \mathcal{D}_{new}$}. This larger dataset allows us to train a new model that achieves better classification performance than the old model. A straight forward way to do so, is to finetune the initial pretrained model on the updated data~(old and new data): 
\begin{equation}
\theta_{new\_target} = \argminA_{\theta} \mathcal{L}(\theta | \theta_{pre}, \mathcal{D}_{upd}).
\label{equ:target}
\end{equation}
This is the process depicted in Figure \ref{fig:motivation} and, unfortunately, leads to high negative flip rate which in turn limits the compatibility between the old and new model. The goal of this work is to find a method that emits a new model which produces minimal negative flips while achieving the same classification performance as the target model:
\begin{equation}
\begin{split}
&\theta^* = \argminA_{\theta} \mathcal{R}(\theta, \theta_{old})\\
&s.t. \; \mathcal{M}(\theta^*) \approx \mathcal{M}(\theta_{new\_target}),
\end{split}
\label{equ:problem}
\end{equation}
where $\mathcal{R}$ is the regression metric and $\mathcal{M}$ measures the classification performance.
To account for variance, the equality of classification metrics can be defined as e.g. overlapping confidence intervals.

\section{Proposed Method: Backward Compatible Weight Interpolation}
\label{sec:proposed}
We start with the intuitive observation that negative flips are reduced when using the old model as the starting point for finetuning the new model:
\begin{equation}
\theta_{new} = \argminA_{\theta} \mathcal{L}(\theta | \theta_{old}, \mathcal{D}_{upd}).
\label{equ:new}
\end{equation}
Next we interpolate between the weights of the old and new model:
\begin{equation}
    \theta_{\mathrm{BCWI}} = \alpha \theta_{old} + (1-\alpha) \theta_{new},
\end{equation}
where $\alpha \in [0.0, 1.0]$ is the interpolation parameter and regulates the trade-off between classification performance and negative flip rate. A larger $\alpha$ moves the model closer to the old model, reducing negative flip rate but ultimately sacrifices the improved classification performance. We empirically show that in all but one of the conducted experiments there exists an~$\alpha>0$ that results in a model that achieves the same classification performance as the target model while significantly reducing negative flips. We call this method Backward Compatible Weight Interpolation~(BCWI).

\subsection{FisherBCWI}
\label{sec:fisher}
The interpolation with a single parameter might not be optimal because not every model weight is equally contributing to a model's predictions. The importance of each weight can be quantified by the diagonal of the empirical Fisher information matrix~\cite{Kirkpatrick2017OvercomingCF, Matena2021MergingMW}: 
\begin{equation}
    F_{old} = \frac{1}{c} \sum_{i}^{N} (\nabla_{\theta_{old}}\log p(y_i|x_i))^2,
\end{equation}
where $c$ is a normalization constant and $\nabla_{\theta_{old}}$ is the gradient in respect to the weights of the old model. 
By using \mbox{$F_{old} \in \mathbb{R}^{|\theta_{old}|}$} as the importance factor for each parameter in the old model we get:
\begin{equation}
    \theta_{\mathrm{FisherBCWI}} = \frac{\alpha F_{old} \theta_{old} + (1-\alpha) \theta_{new}}{\alpha F_{old} + (1-\alpha)},
\end{equation}
where all operations are elementwise. The interpolation is focused on weights that are important for the old model and thus minimizes interference with the weights of the new model.

\subsection{SoupBCWI}
\label{sec:soup}
Ensembling the logits of multiple new models reduces negative flips~\cite{Yan2021PositiveCongruentTT, Xie2021RegressionBA}. The inference cost increases linearly with each new model in the ensemble and makes it impracticable for many applications. To alleviate this, we employ a soup ensemble~\cite{Wortsman2022ModelSA} of new models. A soup ensemble is formed by averaging the weights of multiple models that were individually finetuned from the same pretrained model. We find that the soup ensemble of new models is reducing negative flips. This is complementary to BCWI as we show by interpolating the ensemble weights towards the weights of the old model:
\begin{equation}
    \theta_{\mathrm{SoupBCWI}} = \alpha \theta_{old} + (1-\alpha) \frac{1}{M}\sum_j^M \theta_{new_j},
\end{equation}
where $M$ is the number of new models. Each new model is finetuned according to Equation~\ref{equ:new} and each with a different random seed. In the next section, we motivate the data update scenarios that we use to demonstrate the effectiveness of the proposed methods.

\section{Data Update Scenarios}
\label{sec:scenarios}
\begin{table}[t]
\centering
\resizebox{.99\linewidth}{!}{
\begin{tabular}{@{}lrrrcrrrrr@{}}
\toprule
&\multicolumn{3}{c}{\textbf{Add\_Data Scenario}}  & \phantom{} &\multicolumn{4}{c}{\textbf{Add\_Classes Scenario}} \\ 
\midrule
 & \textbf{Train} & \textbf{Dev} & \textbf{Test} & \phantom{}  & \textbf{Train} & \textbf{Dev} & \textbf{Test} & \textbf{\#C}\\
\midrule
\textbf{MASSIVE}\\
old         & 1,000    & 333  & 4,000 && 1,222   & 409       & 3,258 & 47\\
+ new        & 500    & 167   & -     && 278   & 91        & 742   & 13\\
= updated     & 1,500   & 500  & 4,000 && 1,500  & 500       & 4,000  & 60\\
\midrule
\textbf{Banking77} \\
old         & 700    & 233  & 4,000 && 927   & 310       & 3,713 & 70\\
+ new        & 300    & 100   & -     && 73    & 23        & 287   & 7\\
= updated     & 1,000  & 333  & 4,000 && 1,000   & 333       & 4,000 & 77\\
\midrule
\textbf{AG News}\\
old     & 120    & 60  & 4,000 && 225  & 113       & 3,000 & 3\\
+ new    & 180    & 90  & -     && 75   & 37       & 1,000 & 1\\
= updated & 300    & 150 & 4,000 && 300  & 150       & 4,000 & 4\\
\bottomrule
\end{tabular}
}
\caption{The dataset splits for the Add\_Data and Add\_Classes update scenarios constructed for MASSIVE~\cite{FitzGerald2022MASSIVEA1}, Banking77~\cite{Casanueva2020EfficientID} and AG News~\cite{Zhang2015CharacterlevelCN}. The \textit{updated}~data is the \textit{old}~data in addition to the \textit{new}~data. The test set is only updated when new classes are added. The~\#C-column lists the number of classes in the respective data portion of the AC~scenario. The AD~scenario includes all classes.}
\label{tab:data}
\end{table}
The data that is available to train a given classification model changes over time. This can be due to several reasons. More labeled data for the existing classes is obtained by annotating instances from the initial source or from observed queries. Data for new classes is added to support additional downstream features or classes are split up to allow for more fine-grained classification.
The retraining of an existing model on the evolved data basis is called \textit{data update}. In this work, we focus on two isolated data update scenarios that cover two common use cases, namely adding i.i.d. data and adding new classes. We simulate the two scenarios in order to study the prevalence and mitigation of regression during data updates.

\subsection{Add\_Data Scenario}
In the Add\_Data (AD) scenario, the amount of available data is increased by adding new instances for the current set of classes. This is the most basic type of data update and aims at improving the classification performance of the derived model. The additional data is usually obtained by annotating more instances from the initial data source or from the observed model queries. While in the latter case the distribution can shift over time, we assume i.i.d. data for this scenario.
\subsection{Add\_Classes Scenario}
In the Add\_Classes (AC) scenario, we study data updates that consists of adding new classes and corresponding instances to the existing data. This is necessary when the text classification based system supports new features. For example, a virtual assistant is extended with a food delivery feature, a news classification model covers emerging topics or medical reports are classified according to new diseases codes. 
\begin{figure*}[ht]
    \centering
    \includegraphics[width=0.98\textwidth]{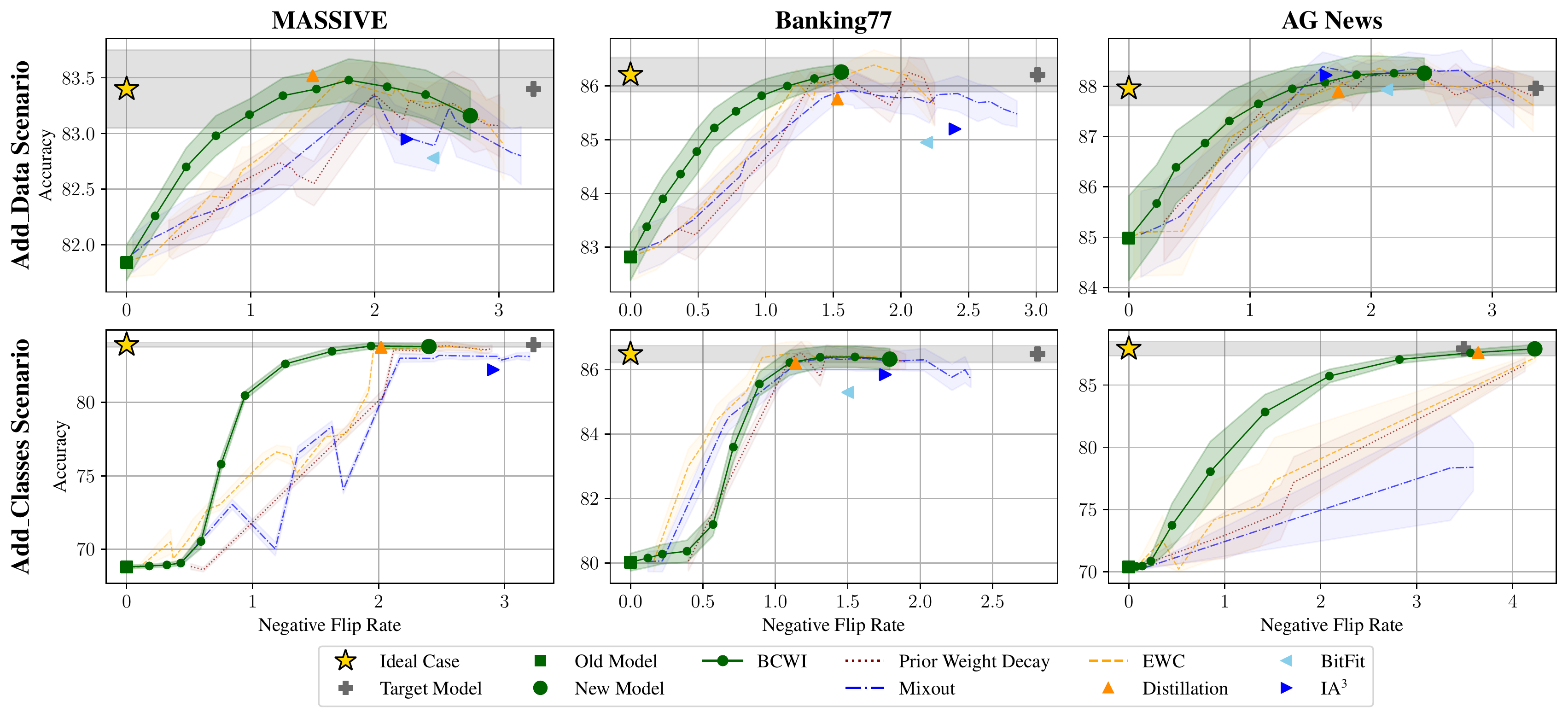}
    \caption{Results for the AD and AC scenario evaluated on our test sets of MASSIVE~\cite{FitzGerald2022MASSIVEA1}, Banking77~\cite{Casanueva2020EfficientID} and AG~News~\cite{Zhang2015CharacterlevelCN}. The gray horizontal bar is spanned by the 95\% confidence interval of the target model and indicates the the level of accuracy a model should reach. Baselines are Prior Weight Decay~\cite{Lee2020MixoutER}, Mixout~\cite{Lee2020MixoutER}, EWC~\cite{Kirkpatrick2017OvercomingCF}, Distillation~\cite{Xie2021RegressionBA}, BitFit~\cite{BenZaken2022BitFitSP} and IA\textsuperscript{3}~\cite{Liu2022FewShotPF}. Identical markers belong to the same method evaluated with different trade-off parameters. For BCWI this is $\alpha$ in 0.1 steps where 0.0 is equivalent to the new model and 1.0 is equivalent to the old model. The trade-off parameters for the baselines are listed in Appendix A. The ideal case for a new model is to have zero negative flips while maintaining the target accuracy. BCWI consistently produces a model that is closer to the ideal case than any of the baseline methods and is more stable across different trade-off parameters.}
    \label{fig:main_results}
\end{figure*}
\subsection{Datasets and Splits}
We simulate the two described data update scenarios for three datasets each. MASSIVE~\cite{FitzGerald2022MASSIVEA1} is a natural language understanding dataset with 60 intents covering basic domains of a virtual assistant. We use the English portion of the data. Banking77~\cite{Casanueva2020EfficientID} includes utterances with 77 intents for a virtual assistant limited to the banking domain. AG~News~\cite{Zhang2015CharacterlevelCN} is a document classification dataset that categorizes news articles into four topics. Table~\ref{tab:data} lists the number of instances in the data splits for both scenarios across all three datasets. We only use a subset of the original data and randomly sample all splits from the training set of the respective dataset. The size of the splits was chosen such that the data update leads to a significant improvement of classification accuracy. In order to simulate the addition of new classes for the AC scenario, we limit its old data splits to a subset of the available classes.

\section{Experiments}
We evaluate our proposed BCWI method on the above described data update scenarios, each constructed for three different datasets.\footnote{Data splits and code to reproduce the experiments can be found at: \url{https://github.com/amazon-science/regression-constraint-model-upgrade/tree/main/nlp}} We choose RoBERTa~\cite{Liu2019RoBERTaAR} as a pretrained model because it is widely used and a representative encoder-only transformer model. The old model and target model are trained by finetuning \robertabase on the old and updated data respectively (see Equation~\ref{equ:old} and~\ref{equ:target}). The new model is trained by finetuning the old model on the updated data~(see Equation~\ref{equ:new}). The empirical Fisher information matrix used by FisherBCWI and EWC is calculated on the training and development instances. The classification performance of each model is reported as accuracy on the updated test set. Regression is measured as negative flip rate~(see Equation~\ref{equ:nfr}) in respect to the classifications of the old model on the updated test set. Experiments are repeated ten times with different random seeds and we report the mean and 95\% confidence interval. Detailed setup and optimization hyperparameters can be found in Appendix A.

\subsection{Baselines}
Because we are the first to explicitly consider the problem of regression during data updates, there are no previous results for this task. Hence we compare BCWI to methods that were introduced in different context but can potentially reduce negative flips during data updates. \citet{Xie2021RegressionBA} propose to use knowledge distillation in order to align the prediction behavior of the old and new model when updating the model architecture. Another group of methods that we compare to are methods to avoid catastrophic forgetting in continual learning. Prior weight decay~\cite{Wiese2017NeuralDA, Lee2020MixoutER} moves the current weights towards the old model at each training step. Mixout~\cite{Lee2020MixoutER} randomly replaces a subset of the weights at each training step with the weights of the old model. \citet{Kirkpatrick2017OvercomingCF} introduce elastic weight consolidation~(EWC) that uses the diagonal Fisher information matrix to weigh the importance of each model parameter in L2 regularization.
BitFit~\cite{BenZaken2022BitFitSP} is a parameter efficient finetuning method that only touches the bias terms of a model. IA\textsuperscript{3}~\cite{Liu2022FewShotPF} introduces additional parameters to scale the outputs of the key and value layer in multi-head attention as well as the position-wise feed-forward
networks. The proper model weights are frozen. All baselines are trained by finetuning the old model according to Equation~\ref{equ:new}.

\subsection{Results}
\label{sec:results}
We first discuss the results for the AD scenario shown in the top row of Figure~\ref{fig:main_results}. Looking at the MASSIVE plot, we see that the new model~(see Equation~\ref{equ:new}) yields lower NFR than the target model~(see Equation~\ref{equ:target}) while achieving similar accuracy. The horizontal gray bar is spanned by the 95\%~confidence interval around the accuracy of the target model and indicates the area of accuracy that fulfills the constraint in Equation~\ref{equ:problem}. The dots along the green line are BCWI models evaluated at decreasing $\alpha$-values with step size 0.1, starting from~$\alpha$=1.0 which is equivalent to the old model on the bottom left to $\alpha$=0.0 which is equivalent to the new model. For all three datasets there is a BCWI model that lies within the gray area and has lower negative flip rate than the new model. The weight regularization baselines are evaluated with different regularization strength and the individual markers for Prior Weight Decay, Mixout and EWC are connected. The plots reveal that the baselines are competitive at accuracy levels close to the new model but drop faster than BCWI when approaching low negative flip rate. The numerical results in Table~\ref{tab:results_add_data} show that BCWI can reduce negative flips by up to three times over the target model while maintaining the accuracy.
\begin{table}[t]
\centering
\resizebox{.985\linewidth}{!}{
\begin{tabular}{@{}lc@{}cc@{}cc@{}c@{}}
\toprule
& \multicolumn{2}{c}{\textbf{MASSIVE}}  & \multicolumn{2}{c}{\textbf{Banking77}} & \multicolumn{2}{c}{\textbf{AG News}}\\ 
\midrule
\textbf{Model} & \textbf{ACC}\textuparrow &  \textbf{NFR}\textdownarrow  & \textbf{ACC}\textuparrow & \textbf{NFR}\textdownarrow  & \textbf{ACC}\textuparrow & \textbf{NFR}\textdownarrow\\ 
\midrule
Old Model     & 81.8 \scriptsize{±0.2}    & \enskip 0.0 \scriptsize{±0.0}         & 82.8 \scriptsize{±0.4}    & \enskip 0.0 \scriptsize{±0.0} & 85.0 \scriptsize{±0.8}    & \enskip 0.0 \scriptsize{±0.0}\\
\midrule
Target Model   & 83.4 \scriptsize{±0.4}    & \enskip 3.3 \scriptsize{±0.4}       & 86.2 \scriptsize{±0.4}    & \enskip 3.0 \scriptsize{±0.3} & 88.0 \scriptsize{±0.1}    & \enskip 3.4 \scriptsize{±0.3}\\
New Model   & 83.2 \scriptsize{±0.2}    & \enskip 2.8 \scriptsize{±0.2}       & 86.3 \scriptsize{±0.1}    & \enskip 1.6 \scriptsize{±0.1} & 88.3 \scriptsize{±0.3}    & \enskip 2.4 \scriptsize{±0.3}\\
\midrule
BitFit   & 82.8 \scriptsize{±0.3}    & \enskip 2.5 \scriptsize{±0.2}  & 85.0\textsuperscript{*} \scriptsize{±0.4}    & \enskip 2.2 \scriptsize{±0.2} & 87.9 \scriptsize{±0.3} & \enskip 2.1 \scriptsize{±0.1}\\
IA\textsuperscript{3}   & 83.0 \scriptsize{±0.2}    & \enskip 2.3 \scriptsize{±0.1}       & 85.2\textsuperscript{*} \scriptsize{±0.4}    & \enskip 2.4 \scriptsize{±0.2} & 88.2 \scriptsize{±0.5} & \enskip \textbf{1.6} \scriptsize{±0.2}\\
Distillation   & 83.5 \scriptsize{±0.2}    & \enskip \textbf{1.5} \scriptsize{±0.2}       & 85.8 \scriptsize{±0.2}    & \enskip 1.5 \scriptsize{±0.2} & 87.9 \scriptsize{±0.5} & \enskip \textbf{1.7} \scriptsize{±0.3}\\
PriorWD   & 83.4 \scriptsize{±0.3}    & \enskip 2.0 \scriptsize{±0.2}       & 85.8 \scriptsize{±0.3}    & \enskip 1.3 \scriptsize{±0.1} & 88.1 \scriptsize{±0.4}    & \enskip \textbf{1.7} \scriptsize{±0.2}\\
Mixout   & 83.0 \scriptsize{±0.2}    & \enskip 1.8 \scriptsize{±0.2}       & 85.8 \scriptsize{±0.3}    & \enskip 1.4 \scriptsize{±0.1} & 88.4 \scriptsize{±0.4} & \enskip \textbf{1.6} \scriptsize{±0.2}\\
EWC   & 83.3 \scriptsize{±0.1}    & \enskip \textbf{1.6} \scriptsize{±0.1}       & 86.1 \scriptsize{±0.2}    & \enskip 1.2 \scriptsize{±0.1} & 87.9 \scriptsize{±0.4} & \enskip \textbf{1.6} \scriptsize{±0.3}\\
\midrule
BCWI   & 83.4 \scriptsize{±0.1}    & \enskip \textbf{1.4} \scriptsize{±0.1}       & 85.5 \scriptsize{±0.3}    & \enskip \textbf{0.8} \scriptsize{±0.1} & 88.0 \scriptsize{±0.4}    & \enskip \textbf{1.5} \scriptsize{±0.2}\\
\bottomrule
\end{tabular}
}
\caption{Add\_Data scenario results for BCWI in comparison to baselines. Hyperparameters are tuned on the dev set. '\textsuperscript{*}' indicates that there is no overlap with the target accuracy. \textbf{Bold} NFR values have overlapping 95\% confidence intervals with the best value~(except old model).}
\label{tab:results_add_data}
\end{table}
\begin{table}[t]
\centering
\resizebox{.999\linewidth}{!}{
\begin{tabular}{@{}lc@{}cc@{}cc@{}c@{}}
\toprule
& \multicolumn{2}{c}{\textbf{MASSIVE}}  & \multicolumn{2}{c}{\textbf{Banking77}} & \multicolumn{2}{c}{\textbf{AG News}}\\ 
\midrule
\textbf{Model} & \textbf{ACC}\textuparrow & \textbf{NFR}\textdownarrow  & \textbf{ACC}\textuparrow & \textbf{NFR}\textdownarrow  & \textbf{ACC}\textuparrow & \textbf{NFR}\textdownarrow\\ 
\midrule
Old Model     & 68.8 \scriptsize{±0.1}    & \enskip 0.0 \scriptsize{±0.0}         & 80.0 \scriptsize{±0.3}    & \enskip 0.0 \scriptsize{±0.0} & 70.4 \scriptsize{±0.1}    & \enskip 0.0 \scriptsize{±0.0}\\
\midrule
Target Model   & 83.9 \scriptsize{±0.2}    & \enskip 3.2 \scriptsize{±0.2}       & 86.5 \scriptsize{±0.3}    & \enskip 2.8 \scriptsize{±0.3} & 87.9 \scriptsize{±0.5}    & \enskip \textbf{3.5} \scriptsize{±0.6}\\
New Model   & 83.8 \scriptsize{±0.2}    & \enskip 2.4 \scriptsize{±0.1}       & 86.3 \scriptsize{±0.3}    & \enskip 1.8 \scriptsize{±0.2} & 87.9 \scriptsize{±0.3}    & \enskip \textbf{4.2} \scriptsize{±0.5}\\
\midrule
BitFit   & 81.6\textsuperscript{*} \scriptsize{±0.2}    & \enskip 3.3 \scriptsize{±0.2}       & 85.3\textsuperscript{*} \scriptsize{±0.4}    & \enskip 1.5 \scriptsize{±0.2} & 86.8\textsuperscript{*} \scriptsize{±0.3} & \enskip 4.9 \scriptsize{±0.7}\\
IA\textsuperscript{3}   & 82.2\textsuperscript{*} \scriptsize{±0.4}    & \enskip 2.9 \scriptsize{±0.2}       & 85.8 \scriptsize{±0.4}    & \enskip 1.8 \scriptsize{±0.2} & 87.1 \scriptsize{±0.3} & \enskip 4.9 \scriptsize{±0.3}\\
Distillation   & 83.8 \scriptsize{±0.2}   & \enskip 2.0 \scriptsize{±0.2}        & 86.2 \scriptsize{±0.3}    & \enskip \textbf{1.1} \scriptsize{±0.1} & 87.6 \scriptsize{±0.2}    & \enskip \textbf{3.6} \scriptsize{±0.5}\\
PriorWD   & 83.3\textsuperscript{*} \scriptsize{±0.3}    & \enskip 2.1 \scriptsize{±0.2}       & 86.3 \scriptsize{±0.3}    & \enskip \textbf{1.1} \scriptsize{±0.1} & 87.4 \scriptsize{±0.4}    & \enskip \textbf{4.3} \scriptsize{±0.4}\\
Mixout   & 83.0\textsuperscript{*} \scriptsize{±0.2}   & \enskip 2.4 \scriptsize{±0.1}        & 86.2 \scriptsize{±0.3}    & \enskip 1.2 \scriptsize{±0.1} & 87.6 \scriptsize{±0.4}    & \enskip 5.0 \scriptsize{±0.5}\\
EWC   & 83.6 \scriptsize{±0.3}   & \enskip 2.0 \scriptsize{±0.1}        & 86.4 \scriptsize{±0.3}    & \enskip \textbf{0.9} \scriptsize{±0.1} & 87.9 \scriptsize{±0.4}    & \enskip \textbf{4.3} \scriptsize{±0.4}\\
\midrule
BCWI   & 83.2\textsuperscript{*} \scriptsize{±0.2}    & \enskip \textbf{1.4} \scriptsize{±0.1}       & 86.0 \scriptsize{±0.4}    & \enskip \textbf{1.0} \scriptsize{±0.1} & 87.6 \scriptsize{±0.3}    & \enskip \textbf{3.6} \scriptsize{±0.4}\\
\bottomrule
\end{tabular}
}
\caption{Add\_Classes scenario results for BCWI in comparison to baselines. Hyperparameters are tuned on the dev set. '\textsuperscript{*}' indicates that there is no overlap with the target accuracy. \textbf{Bold} NFR values have overlapping 95\% confidence intervals with the best value~(except old model).}
\label{tab:results_add_classes}
\end{table}

The second row in Figure~\ref{fig:main_results} features the BCWI and baseline plots for the AC scenario. The green line which connects individual BCWI models follows an S-shaped curve with an inflection point near $\alpha \geq$~0.5 (on AG~News the lower end of the S is compressed along the x-axis). This is because the old model is not trained on the new classes and their accuracy drops rapidly to zero once the old model weights dominate. For each dataset in the AC scenario there is a BCWI model that lies within the target accuracy and yield lower NFR than the new model. On AG~News none of the $\alpha$-values within the gray area result in lower NFR than the target model. The numerical values in Table~\ref{tab:results_add_classes} show that BCWI is as good as or better than the baselines in reducing regression at the same accuracy level.

\subsection{BCWI Variants}
\begin{figure}[t]
    \centering
    \begin{subfigure}{0.99\linewidth}
    \includegraphics[width=0.99\linewidth]{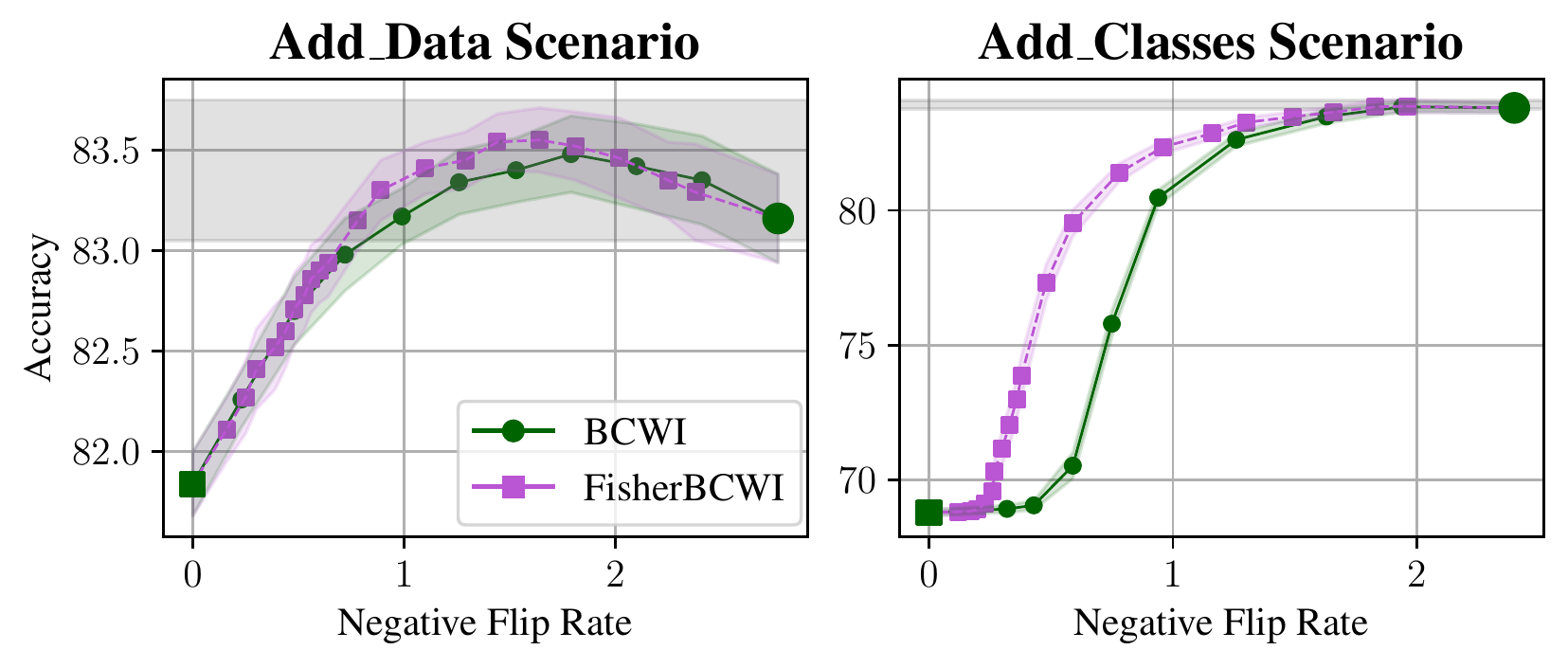}
    \end{subfigure}
    \begin{subfigure}{0.99\linewidth}
    \includegraphics[width=0.99\linewidth]{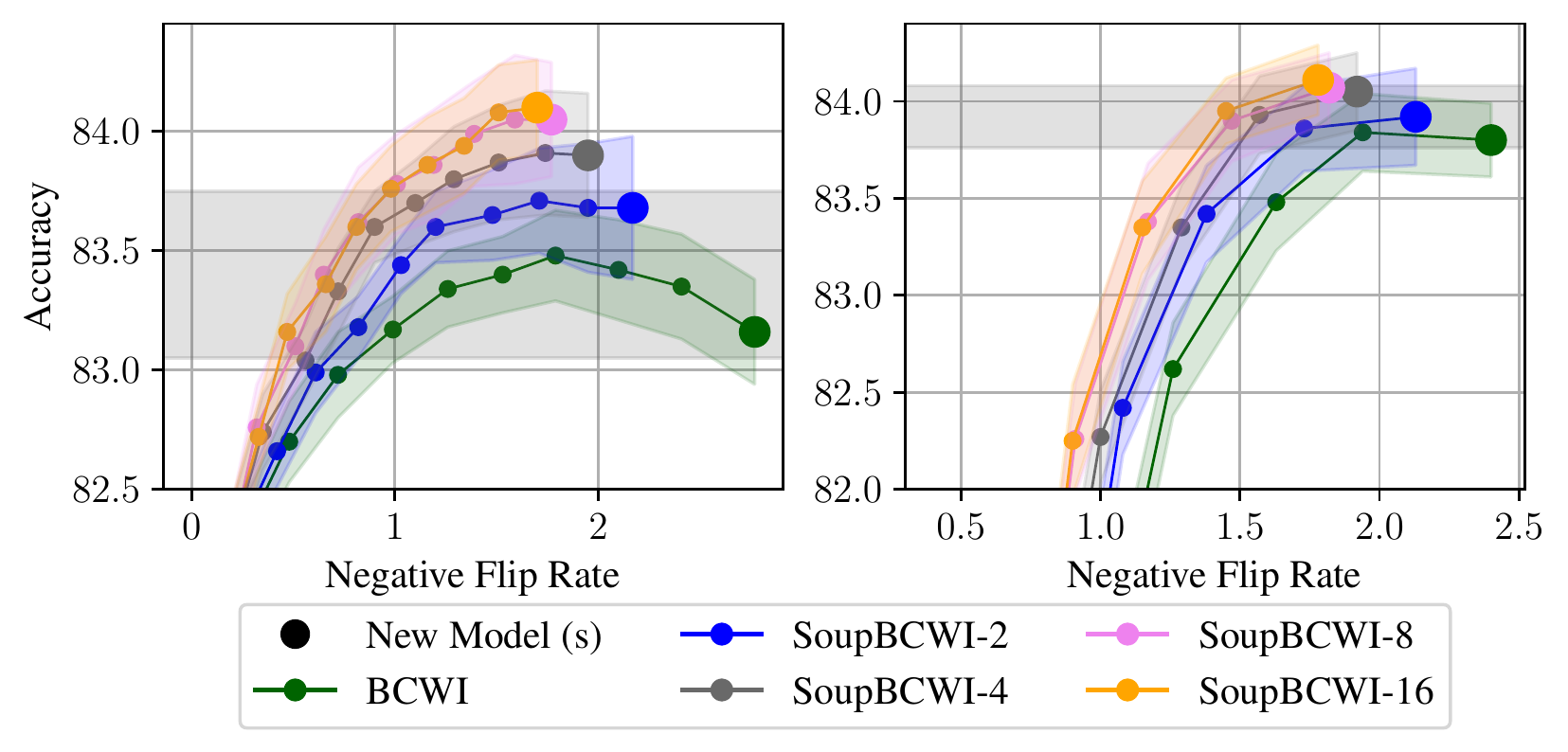}
    \end{subfigure}
    \caption{Plots for FisherBCWI~(top), SoupBCWI~(bottom) in comparison with vanilla BCWI on MASSIVE. Add\_Data scenario~(left) and Add\_Classes scenario~(right). Models within the gray area maintain the target accuracy. FisherBCWI uses the Fisher information matrix as importance weighting during interpolation. SoupBCWI-M is a soup ensemble~\cite{Wortsman2022ModelSA} with M new model and interpolation towards the old model.}
    \label{fig:variants_massive}
\end{figure}
\label{sec:results_variants}
\begin{table}[t]
\centering
\resizebox{.999\linewidth}{!}{
\begin{tabular}{@{}lc@{}cc@{}cc@{}c@{}}
\toprule
& \multicolumn{2}{c}{\textbf{MASSIVE}}  & \multicolumn{2}{c}{\textbf{Banking77}} & \multicolumn{2}{c}{\textbf{AG News}}\\ 
\midrule
\textbf{Model} & \textbf{ACC} &  \textbf{NFR}\textdownarrow  & \textbf{ACC} & \textbf{NFR}\textdownarrow  & \textbf{ACC} & \textbf{NFR}\textdownarrow\\ 
\midrule
\multicolumn{7}{c}{\textbf{Add\_Data Scenario}} \\
\midrule
BCWI   & 83.4 \scriptsize{±0.1}    & \enskip 1.4 \scriptsize{±0.1}       & 85.5 \scriptsize{±0.3}    & \enskip 0.8 \scriptsize{±0.1} & 88.0 \scriptsize{±0.4}    & \enskip 1.5 \scriptsize{±0.2}\\
\midrule
FisherBCWI   & 83.5 \scriptsize{±0.2}    & \enskip 2.0 \scriptsize{±0.2}       & 85.5 \scriptsize{±0.3}    & \enskip 0.6 \scriptsize{±0.1} & 88.1 \scriptsize{±0.4}    & \enskip 1.9 \scriptsize{±0.3}\\
\midrule
SoupBCWI-2   & 83.5 \scriptsize{±0.1}    & \enskip \textbf{1.1} \scriptsize{±0.1}       & 85.6 \scriptsize{±0.4}    & \enskip 0.7 \scriptsize{±0.1} & 88.0 \scriptsize{±0.4}    & \enskip 1.2 \scriptsize{±0.2}\\
SoupBCWI-4   & 83.6 \scriptsize{±0.1}    & \enskip \textbf{0.9} \scriptsize{±0.1}       & 85.4 \scriptsize{±0.4}    & \enskip 0.6 \scriptsize{±0.1} & 88.0 \scriptsize{±0.3}    & \enskip \textbf{1.1} \scriptsize{±0.1}\\
SoupBCWI-8   & 83.6 \scriptsize{±0.2}    & \enskip \textbf{0.8} \scriptsize{±0.1}       & 85.4 \scriptsize{±0.3}    & \enskip 0.6 \scriptsize{±0.1} & 88.0 \scriptsize{±0.3}    & \enskip \textbf{1.1} \scriptsize{±0.1}\\
SoupBCWI-16   & 83.5 \scriptsize{±0.2}    & \enskip \textbf{0.7} \scriptsize{±0.1}       & 85.4 \scriptsize{±0.4}    & \enskip 0.6 \scriptsize{±0.1} & 87.9 \scriptsize{±0.4}    & \enskip \textbf{0.9} \scriptsize{±0.1}\\
\midrule
\multicolumn{7}{c}{\textbf{Add\_Classes Scenario}} \\
\midrule
BCWI   & 83.2 \scriptsize{±0.2}    & \enskip 1.4 \scriptsize{±0.1}       & 86.0 \scriptsize{±0.4}    & \enskip 1.0 \scriptsize{±0.1} & 87.6 \scriptsize{±0.3}    & \enskip 3.6 \scriptsize{±0.4}\\
\midrule
FisherBCWI   & 82.9 \scriptsize{±0.2}    & \enskip 1.2 \scriptsize{±0.1}       & 85.7 \scriptsize{±0.5}    & \enskip \textbf{0.7} \scriptsize{±0.1} & 87.5 \scriptsize{±0.2}    & \enskip 3.3 \scriptsize{±0.4}\\
\midrule
SoupBCWI-2   & 83.0 \scriptsize{±0.3}    & \enskip 1.2 \scriptsize{±0.1}       & 85.8 \scriptsize{±0.4}    & \enskip 0.8 \scriptsize{±0.1} & 87.9 \scriptsize{±0.3}    & \enskip 3.8 \scriptsize{±0.3}\\
SoupBCWI-4   & 82.9 \scriptsize{±0.2}    & \enskip \textbf{1.1} \scriptsize{±0.1}       & 85.8 \scriptsize{±0.3}    & \enskip \textbf{0.6} \scriptsize{±0.1} & 87.9 \scriptsize{±0.3}    & \enskip 3.8 \scriptsize{±0.4}\\
SoupBCWI-8   & 82.9 \scriptsize{±0.3}    & \enskip \textbf{1.0} \scriptsize{±0.1}       & 85.8 \scriptsize{±0.3}    & \enskip \textbf{0.5} \scriptsize{±0.1} & 87.7 \scriptsize{±0.3}    & \enskip 3.5 \scriptsize{±0.3}\\
SoupBCWI-16   & 82.9 \scriptsize{±0.2}    & \enskip \textbf{1.0} \scriptsize{±0.1}       & 85.8 \scriptsize{±0.3}    & \enskip \textbf{0.5} \scriptsize{±0.1} & 87.9 \scriptsize{±0.3}    & \enskip 3.8 \scriptsize{±0.3}\\
\bottomrule
\end{tabular}
}
\caption{Results for FisherBCWI and SoupBCWI in comparison with BCWI. \textbf{Bold} NFR values are lower than those of BCWI and without overlapping 95\% confidence intervals.}
\label{tab:results_variants}
\end{table}
We discuss the results for the BCWI variants proposed in Section~\ref{sec:fisher} and~\ref{sec:soup} in this section. FisherBCWI uses the diagonal Fisher information matrix as importance weighting when interpolating between old and new model. In Figure~\ref{fig:variants_massive}~(top), we can see that its trade-off trajectory is slightly favorable to the one of vanilla BCWI, especially in the AC scenario. This shows that studying interpolation schemes beyond linear is a promising research direction to further reduce negative flips. However, within the target accuracy area, there is no significant NFR improvement (see Table~\ref{tab:results_variants}). 

The results for SoupBCWI are presented in Figure ~\ref{fig:variants_massive}~(bottom row). The large dots at the right end of the SoupBCWI graphs represents the soup ensemble of multiple new models without old model interpolation. Their location in the graph reveals that a soup ensemble of multiple new models not only increases accuracy but also reduces regression. The effect slows down after more than four models in the soup. Interpolating the weights of the soup ensemble with the weight of the old model further reduces negative flips. The results in Table~\ref{tab:results_variants} show that with SoupBCWI-4, the negative flip rate is significantly reduced in four out of six experiments in comparison to vanilla BCWI.

\section{Analysis}
\subsection{Method Properties}
\label{sec:properties}
\begin{table}[t]
\centering
\resizebox{.99\linewidth}{!}{
\begin{tabular}{@{}lrrcc}
\toprule
          & \multicolumn{1}{c}{\textbf{Additional}} & \multicolumn{1}{c}{\textbf{Training}} & \multicolumn{1}{c}{\textbf{Tune}} & \multicolumn{1}{c}{\textbf{Inference}}\\
          & \multicolumn{1}{c}{\textbf{Memory}} & \multicolumn{1}{c}{\textbf{Time}} & \multicolumn{1}{c}{\textbf{Trade-Off}} & \multicolumn{1}{c}{\textbf{Cost}}\\
\midrule
\textbf{EWC} & $|F|$ + $|\theta_{old}|$ & $t(F)$ + 1.9x & retrain & 1x \\
\textbf{Prior WD} & $|\theta_{old}|$ & 1.1x & retrain & 1x \\
\textbf{Mixout} & $|\theta_{old}|$ & 1.6x & retrain & 1x \\
\midrule
\textbf{BCWI} & - & 1x & post training & 1x \\
\textbf{FisherBCWI} & - & $t(F)$ + 1x & post training & 1x \\
\textbf{SoupBCWI} & - & $M$x & post training & 1x \\
\midrule
\textbf{Ensemble} & - & 1x & post training & 2x \\
\end{tabular}
}
\caption{Properties of the proposed methods in comparison to considered baselines. \textit{Additional~Memory}:~Number of additional values that need to be held in GPU memory during training. \textit{Training~Time}:~Factor by which the training time of the new model is increased. \textit{Tune~Trade-Off}:~Weather it is necessary to retrain the model in order to tune the accuracy-NFR trade-off. \textit{Inference~Cost}:~Factor by which inference cost is increased. $F$ is the diagonal Fisher information matrix with size $|\theta|$ and with compute time $t(F)$ roughly equal to one epoch. $M$ is the number of new models in the soup ensemble.}
\label{tab:properties}
\end{table}
\begin{figure}[t]
    \centering
    \includegraphics[width=0.99\linewidth]{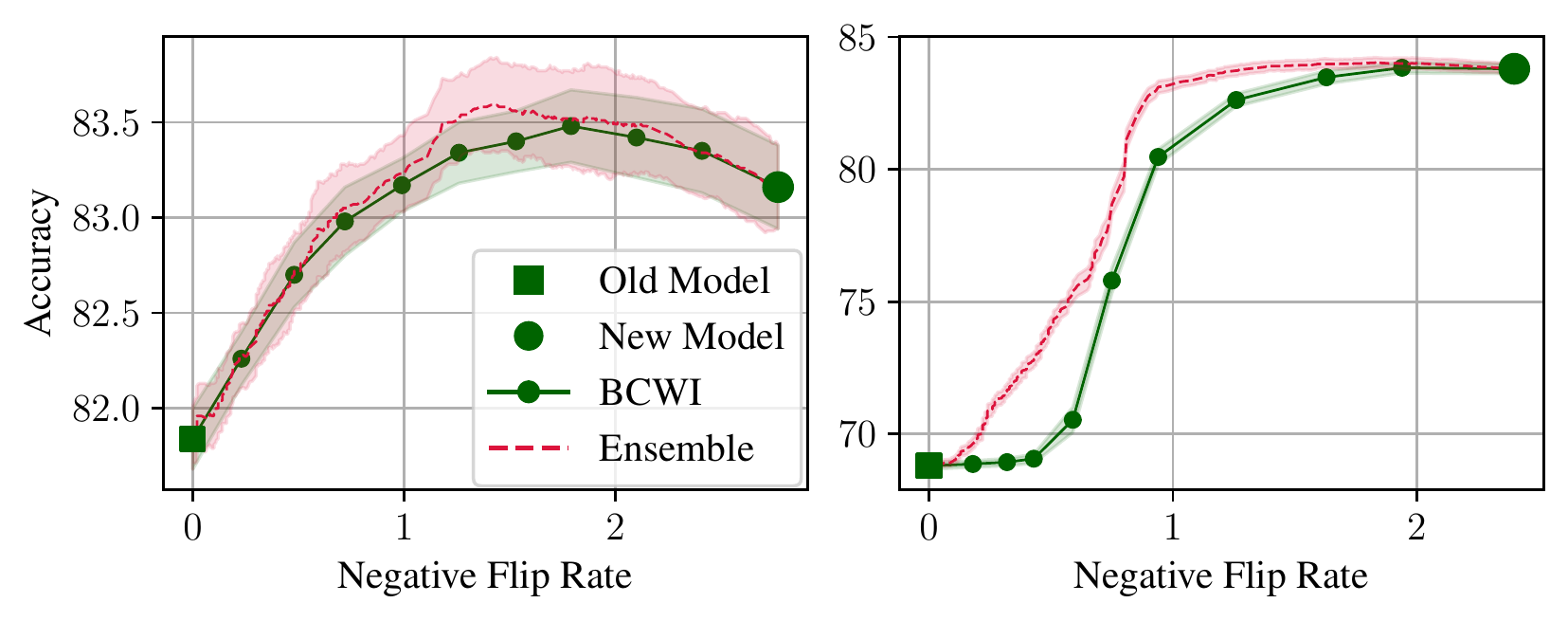}
    \caption{Plots for output ensemble of old and new model in comparison with vanilla BCWI on MASSIV. The ensemble is calculated as the weighted average of output probabilities. AD scenario is on the left and AC scenario on the right.}
    \label{fig:ensemble_massive}
\end{figure}
In this section we discuss the training and inference resources required by BCWI and the utilized baselines listed in Table \ref{tab:properties}. The weight regularization baselines have higher GPU memory requirements because they need to access the weights of the old model at each training step. The calculations necessary to evaluate the regularization terms amount to $1.1-1.9\times$ longer training time. EWC additionally keeps the Fisher information matrix in memory, which is pre-calculated before training. The pre-calculation takes the time of roughly one epoch of training. This is also necessary for the FisherBCWI method. In order to tune the regularization strength of EWC, Prior WD and Mixout, the model needs to be retrained entirely. On the other hand, the $\alpha$-value of BCWI is tuned after training is completed by interpolating the converged model weights. This property of BCWI is a big advantage because it saves training resources and allows to quickly adjust to e.g.~user complaints about too many regression errors in an updated model.

\begin{figure*}[ht]
  \centering
  \begin{subfigure}{0.25\textwidth}
  \includegraphics[width=\textwidth]{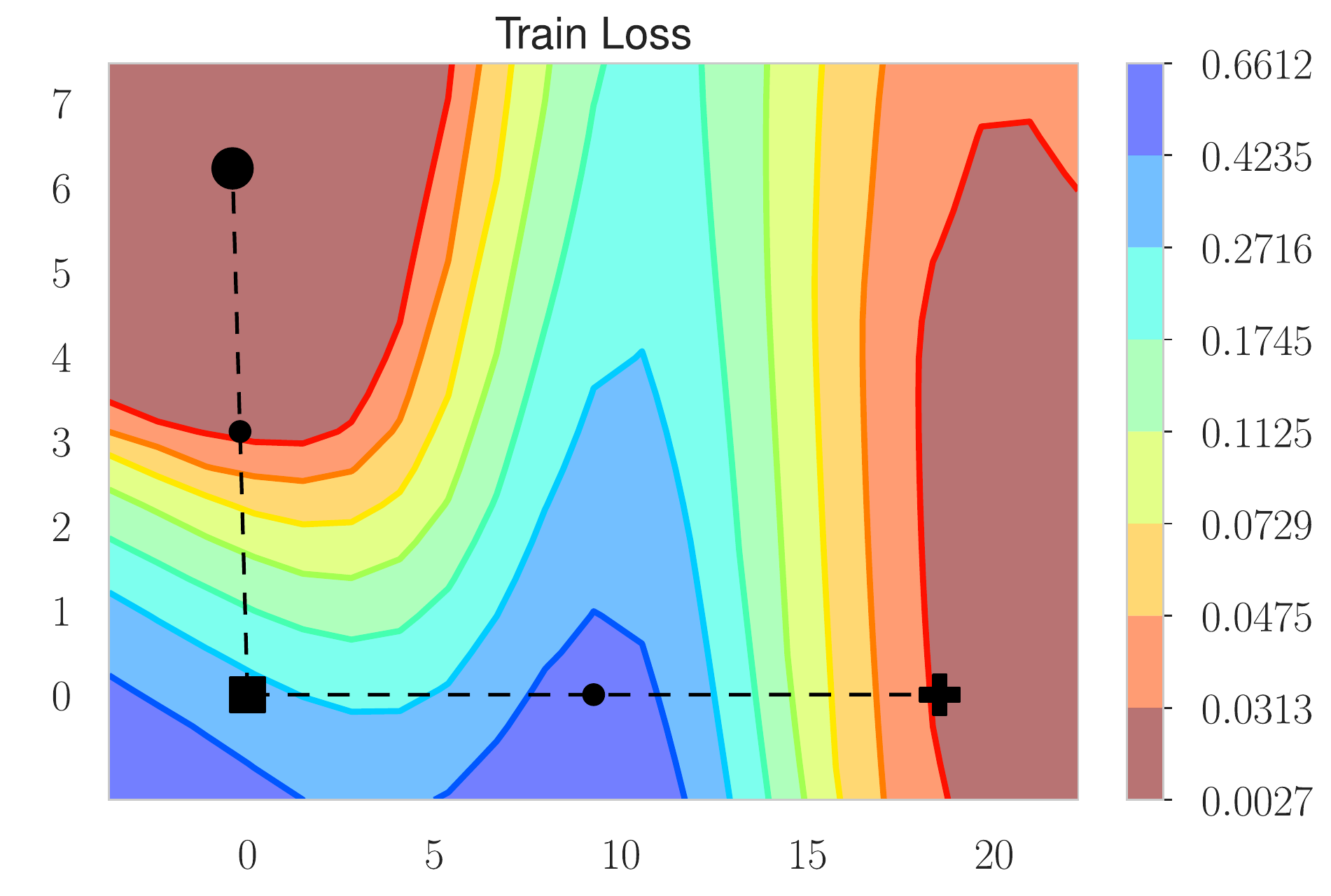}
  \end{subfigure}
  \hspace{0.6cm}
  \begin{subfigure}{0.25\textwidth}
  \includegraphics[width=\textwidth]{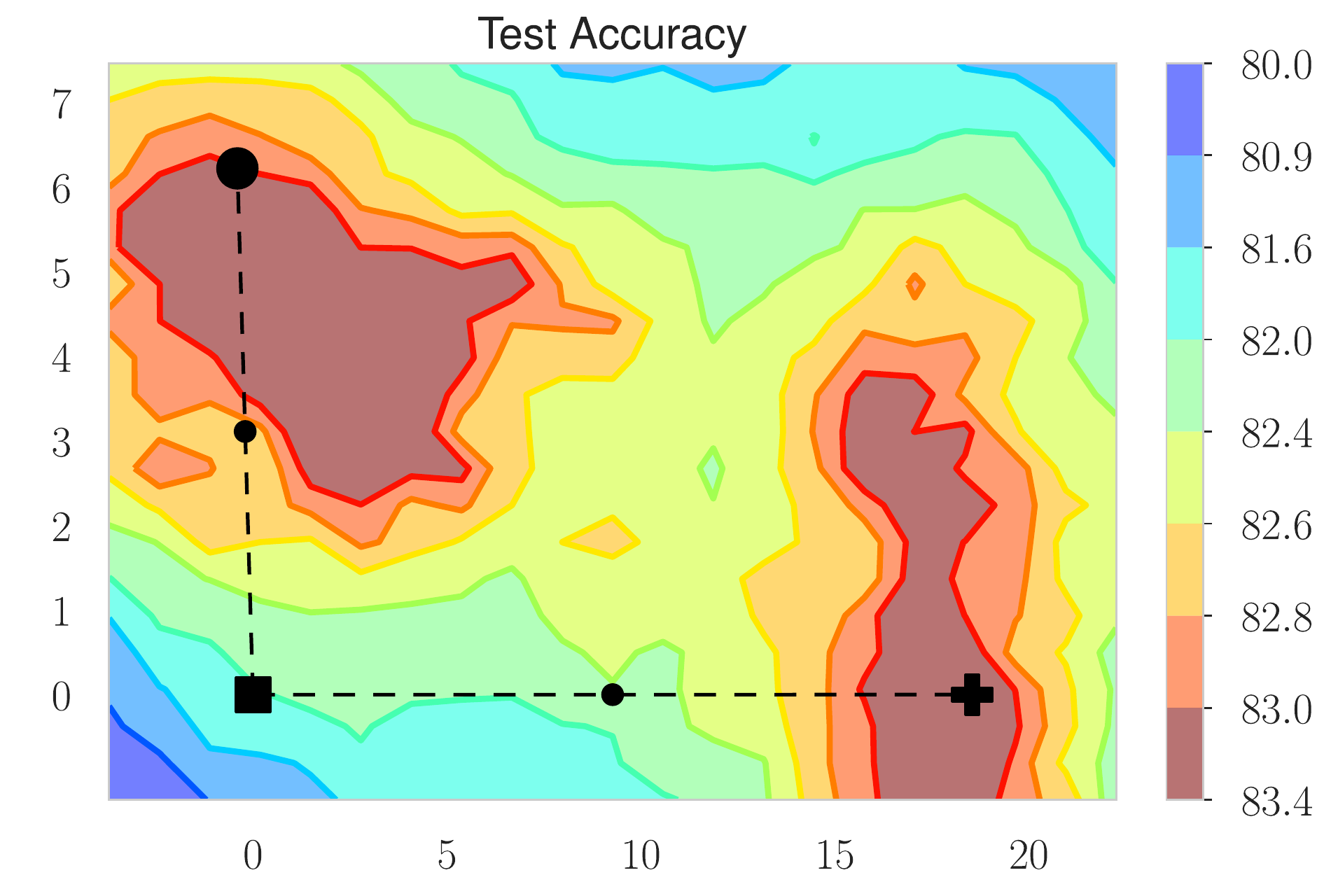}
  \end{subfigure}
  \hspace{0.6cm}
  \begin{subfigure}{0.25\textwidth}
  \includegraphics[width=\textwidth]{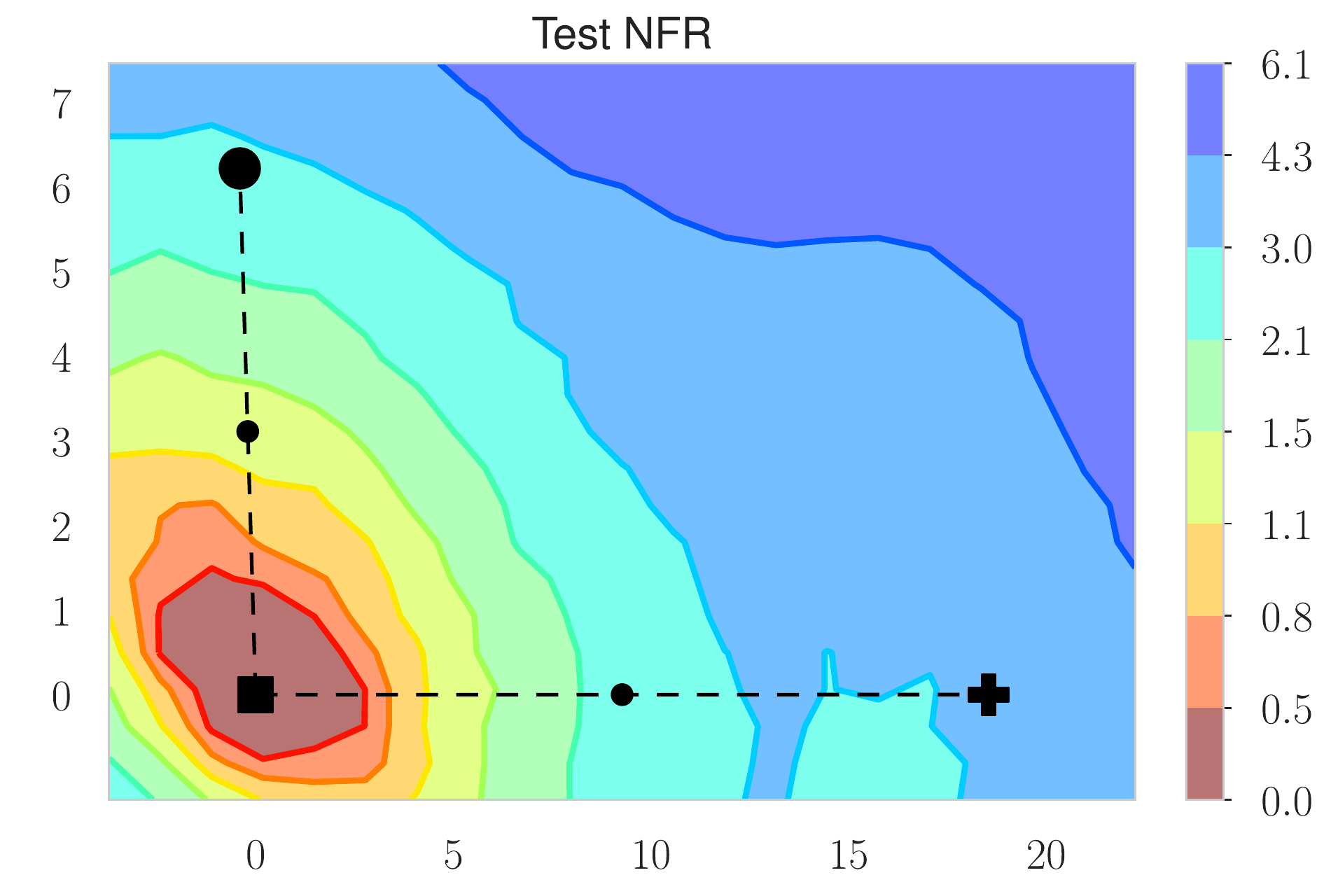}
  \end{subfigure}
  \begin{subfigure}{0.5\textwidth}
  \includegraphics[width=\textwidth]{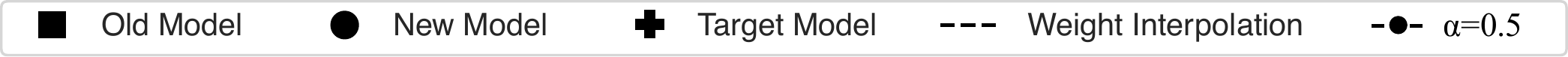}
  \end{subfigure}
  \caption{Visualization of training loss, test accuracy and negative flip rate for the AD scenario on MASSIVE. Visualization technique from~\citet{Izmailov2018AveragingWL}. The x and y axes denote euclidean distance. On the bottom left in each plot is the old model and the dotted lines represent the points along the linear interpolation towards the new model and target model.}
  \label{fig:loss_landscape}
\end{figure*}

\subsection{Output Ensemble of Old and New Model}
\label{sec:ensemble_old_new}
The weight interpolation between old and new model can be seen as an ensemble in weight space. But in contrast to output ensembles, it does not increase inference cost. In an output ensemble, the input needs to be passed through both models and the final prediction is computed from the two output probability distributions. This renders output ensembles impracticable in many applications, especially real-time systems. Figure~\ref{fig:variants_massive}~(bottom) shows the graphs for weight interpolation and weighted average of output probabilities. The trajectories in the AD scenario are similar and in the AC scenario the output ensemble performs slightly better. This highlights that BCWI conveys most of the improvements in regression mitigation, but without the downside of increased inference cost from passing the old and new model.

\subsection{Loss Landscape}
To better understand the dynamics of BCWI, we visualize the loss and error landscapes for the old, new and target model in Figure \ref{fig:loss_landscape}. The left plot shows the cross-entropy loss on the updated training data. The new model and target model, both trained on the updated data, achieve equally low loss. Because the new model is initialized by the old model~(see Equation~\ref{equ:new}), it stays within the same basin. The target model, initialized by the pretrained model~(see Equation~\ref{equ:target}), diverges more from the old model and ends up in a different local minimum. Thus interpolation between the old model and target model faces a high loss barrier and in turn low test accuracy. The distance between old model and target model is three times larger than the distance between old model and new model. According to \citet{rame2022diwa} this leads to a large locality term and makes the models less "averagable". One possible way to alleviate this is permutating the weights~\cite{Ainsworth2022GitRM} of the target model such that it lies within the same basin as the old model. We leave the re-basin of the target model for future work. The plot in the middle shows the accuracy along the interpolation from old to new model and that small $\alpha$-values maintain high accuracy. The interpolation towards the target model traverses low accuracy regions and only achieves high accuracy very close to the target model. The plot on the right shows that the area of low negative flip rate is centered around the old model. This explains the lower NFR for the new model opposed to the target model because the distance between the old and new model is smaller than between the old and target model. Interpolating the weights of the new model and the weights of the old model follows a monotonic decrease of negative flips. This allows to find a point in weight space that has low negative flips while maintaining high accuracy.

\section{Conclusion}
We studied the problem of regression during data updates in text classification. Retraining a model with a larger amount of training data increases accuracy but also introduces negative flips. We propose BCWI which describes the interpolation between the weights of the old model and the weight of the new model. We empirically show on three datasets and two update scenarios that BCWI models significantly reduce negative flips while not sacrificing accuracy. We compare BCWI to strong continual learning methods and achieve similar or better results, while not increasing training or inference cost. Another big advantage of BCWI is that the trade-off parameter $\alpha$ can be tuned without retraining the model. This saves additional training cost and only requires to store the weights of the old and new model. We extend BCWI by using the Fisher information matrix as importance factor in weight interpolation and show that it leads to a favorable trade-off trajectory. Using multiple new models as in proposed SoupBCWI further reduces regression without increasing the inference cost. In principle BCWI is architecture and task agnostic with the possibility to explore applications such as natural language generation left for future work.

\bibliography{example_paper}
\bibliographystyle{icml2023}

\clearpage
\appendix
\section{Experiment Details}
We list the hyperparameters used for training the different models in Table~\ref{tab:hyperparameter}. The selection of hyperparameter largely follows~\citep{Mosbach2021OnTS}. We use the \robertabase model from HuggingFace\footnote{\url{https://huggingface.co/roberta-base}}. The best learning rate and number of epochs is selected on the dev set based on accuracy and NFR. Although there are no extensive experiments, we noticed that BCWI is largely insensitive to hyperparameter selection. The focus can remain on optimizing accuracy and BCWI handles regression after the successful training. The interpolation parameter $\alpha$ is tuned on the dev set by choosing the largest $\alpha$-value that does not cause the accuracy to drop below a chosen threshold (see Table Table~\ref{tab:results_dev_add_data} and~\ref{tab:results_dev_add_classes}). The regularization strength of the baselines is tuned in the same way by selecting the strongest regularization parameter that does not sacrifice accuracy below that threshold on the dev set.

\begin{table}[h]
\centering
\resizebox{.85\linewidth}{!}{
\begin{tabular}{@{}lccc}
\toprule
& \multicolumn{3}{c}{(Range of) Hyperparameters} \\
\midrule
Prior WD & \multicolumn{3}{c}{0.01, 0.1, 1.0, 10.0, 100, 200,}\\
& \multicolumn{3}{c}{1e3, 2e3, 4e3, 1e4, 1e5} \\
Mixout & \multicolumn{3}{c}{0.05, 0.1, 0.2, 0.3, 0.4,}\\
& \multicolumn{3}{c}{0.5, 0.6, 0.7, 0.8, 0.9,}\\
& \multicolumn{3}{c}{0.95, 0.98, 0.99, 0.999} \\
EWC & \multicolumn{3}{c}{1e-5, 1e-4, 1e-3, 0.01,}\\
& \multicolumn{3}{c}{0.1, 1.0, 2.0, 5.0, 10.0, 50.0} \\
& \multicolumn{3}{c}{100, 1e3, 1e4} \\
BitFit \& IA\textsuperscript{3} & \multicolumn{3}{c}{E: 8, 12, 16; LR: 1e-4, 1e-3, 1e-2}\\
\midrule
LR Schedule & \multicolumn{3}{c}{linear} \\
Warmup Ratio & \multicolumn{3}{c}{0.1} \\
Batch Size & \multicolumn{3}{c}{16} \\
Adam $\epsilon$ & \multicolumn{3}{c}{1e-6} \\
Adam $\beta_1$ & \multicolumn{3}{c}{0.9} \\
Adam $\beta_2$ & \multicolumn{3}{c}{0.98} \\
Adam Bias Corr. & \multicolumn{3}{c}{True} \\
Dropout & \multicolumn{3}{c}{0.1} \\
Weight Decay & \multicolumn{3}{c}{0.01} \\
Clip grad. norm & \multicolumn{3}{c}{5.0} \\
\midrule
& \textbf{MASSIVE} & \textbf{Banking77} & \textbf{AG News}\\
\midrule
\textbf{Old Model:} &\\
Epochs & 16 & 16 & 8 \\
Learning Rate & 6e-5 & 6e-5 & 6e-5 \\
\textbf{Target Model:} &\\
Epochs & 16 & 16 & 8 \\
Learning Rate & 6e-5 & 6e-5 & 6e-5 \\
\textbf{New Model:} &\\
Epochs & 3, 6, 10 & 3, 6, 10 & 2, 3, 6 \\
Learning Rate & 3e-5, 6e-5 & 3e-5, 6e-5 & 3e-5, 6e-5 \\

\bottomrule
\end{tabular}
}
\caption{Hyperparameter for the different datasets and methods.}
\label{tab:hyperparameter}
\end{table}

\section{Additional Results}
In Table~\ref{tab:results_dev_add_data} and~\ref{tab:results_dev_add_classes} we show the dev set results for Table~\ref{tab:results_add_data} and~\ref{tab:results_add_classes}. The hyperparameter for the respective method was tuned to reach the accuracy threshold on the dev set. 

\begin{figure}[ht]
    \centering
    \includegraphics[width=0.99\linewidth]{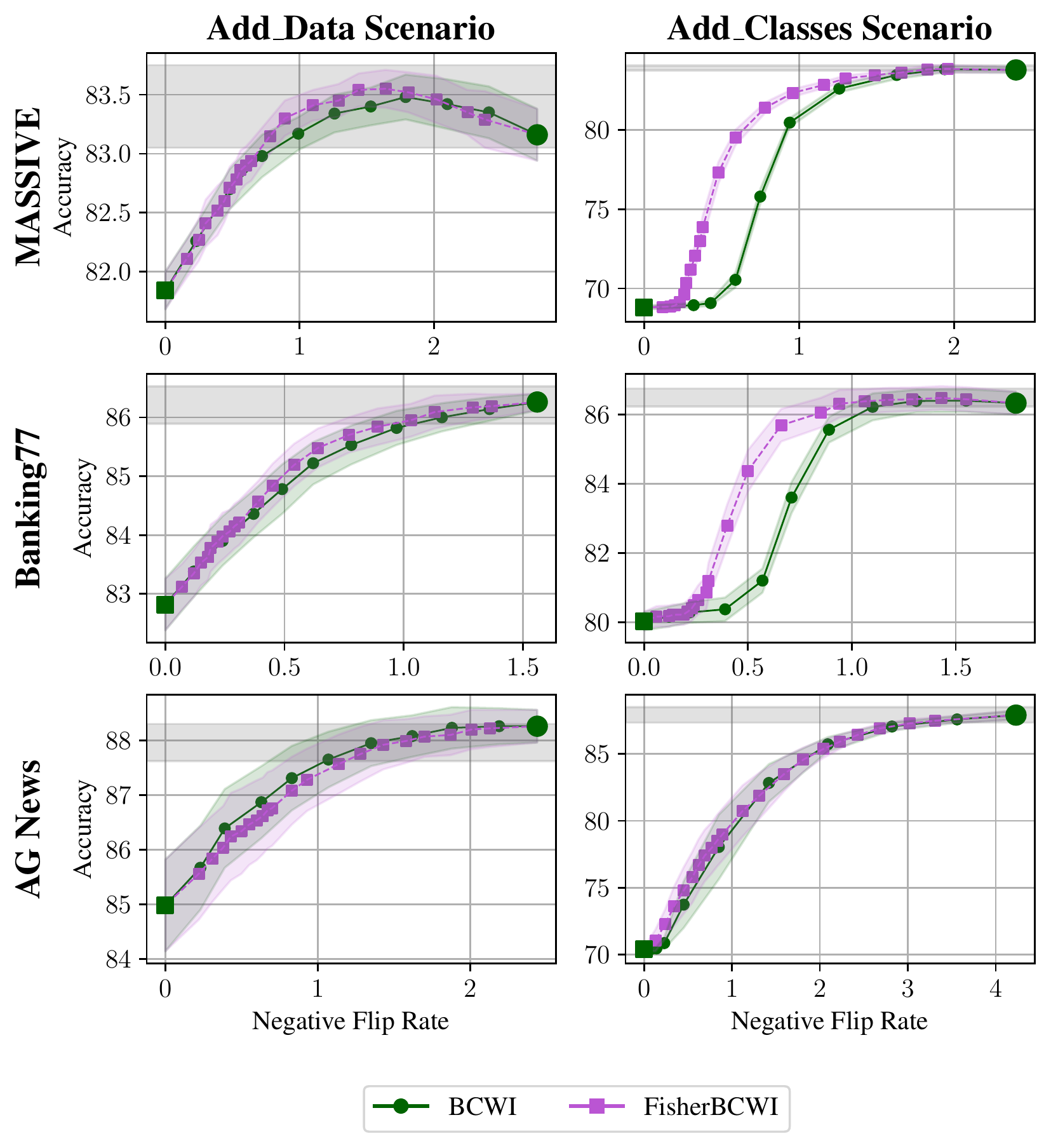}
    \caption{Plots for FisherBCWI in comparison with vanilla BCWI. The gray area indicates the target accuracy level.}
    \label{fig:FisherBCWI}
\end{figure}

We present the plots for FisherBCWI results in Figure~\ref{fig:FisherBCWI}. Results for Soup ensembles and probability ensembles of new models are listed in Table~\ref{tab:results_soup_add_data} and~\ref{tab:results_soup_add_classes}. They achieve the same accuracy and NFR which means that soup ensembles are as good as probability ensembles in reducing regression without increasing inference cost.

\begin{figure}[ht]
    \centering
    \includegraphics[width=0.99\linewidth]{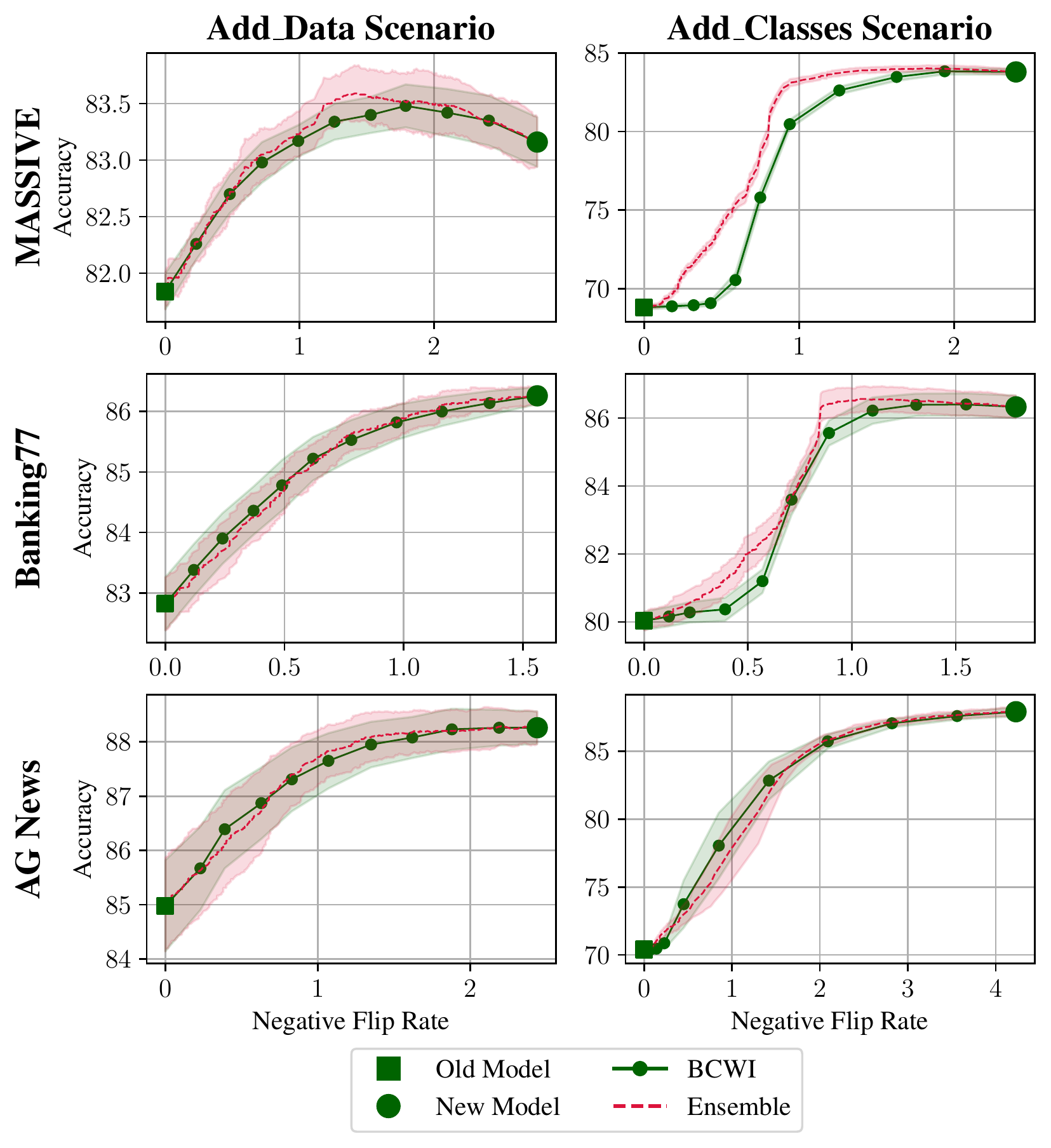}
    \caption{Plots comparing BCWI with the probability ensemble of old and new model.}
    \label{fig:ensemble_other}
\end{figure}

In the analysis in Section~\ref{sec:ensemble_old_new}, we show that trajectory of BCWI closely follows the probability ensemble of old and new model. In Figure~\ref{fig:ensemble_other}. In the AC scenario the probabilities for new classes predicted by the old model are set to zero, because it was only trained on the old classes.
\begin{table}[h]
\centering
\resizebox{.999\linewidth}{!}{
\begin{tabular}{@{}lc@{}cc@{}cc@{}c@{}}
\toprule
& \multicolumn{2}{c}{\textbf{MASSIVE}}  & \multicolumn{2}{c}{\textbf{Banking77}} & \multicolumn{2}{c}{\textbf{AG News}}\\ 
\midrule
\textbf{Model} & \textbf{ACC}\textuparrow &  \textbf{NFR}\textdownarrow  & \textbf{ACC}\textuparrow & \textbf{NFR}\textdownarrow  & \textbf{ACC}\textuparrow & \textbf{NFR}\textdownarrow\\ 
\midrule
Old Model     & 81.8 \scriptsize{±0.2}    & \enskip 0.0 \scriptsize{±0.0}         & 82.8 \scriptsize{±0.4}    & \enskip 0.0 \scriptsize{±0.0} & 85.0 \scriptsize{±0.8}    & \enskip 0.0 \scriptsize{±0.0}\\
\midrule
Target Model   & 83.4 \scriptsize{±0.4}    & \enskip 3.3 \scriptsize{±0.4}       & 86.2 \scriptsize{±0.4}    & \enskip 3.0 \scriptsize{±0.3} & 88.0 \scriptsize{±0.1}    & \enskip 3.4 \scriptsize{±0.3}\\
New Model   & 83.2 \scriptsize{±0.2}    & \enskip 2.8 \scriptsize{±0.2}       & 86.3 \scriptsize{±0.1}    & \enskip 1.6 \scriptsize{±0.1} & 88.3 \scriptsize{±0.3}    & \enskip 2.4 \scriptsize{±0.3}\\
\midrule
Ensemble-2   & 83.8 \scriptsize{±0.3}    & \enskip 2.2 \scriptsize{±0.2}       & 86.4 \scriptsize{±0.2}    & \enskip 1.4 \scriptsize{±0.2} & 88.4 \scriptsize{±0.2}    & \enskip 2.4 \scriptsize{±0.4}\\
Ensemble-4   & 84.0 \scriptsize{±0.2}    & \enskip 2.0 \scriptsize{±0.1}       & 86.5 \scriptsize{±0.2}    & \enskip 1.4 \scriptsize{±0.2} & 88.6 \scriptsize{±0.2}    & \enskip 2.3 \scriptsize{±0.3}\\
Ensemble-8   & 84.2 \scriptsize{±0.2}    & \enskip 1.8 \scriptsize{±0.1}       & 86.5 \scriptsize{±0.2}    & \enskip 1.3 \scriptsize{±0.2} & 88.7 \scriptsize{±0.2}    & \enskip 2.2 \scriptsize{±0.3}\\
Ensemble-16   & 84.3 \scriptsize{±0.2}    & \enskip 1.7 \scriptsize{±0.1}       & 86.4 \scriptsize{±0.2}    & \enskip 1.3 \scriptsize{±0.2} & 88.8 \scriptsize{±0.2}    & \enskip 2.1 \scriptsize{±0.2}\\
\midrule
Soup-2   & 83.7 \scriptsize{±0.3}    & \enskip 2.2 \scriptsize{±0.2}       & 86.3 \scriptsize{±0.2}    & \enskip 1.4 \scriptsize{±0.2} & 88.5 \scriptsize{±0.2}    & \enskip 2.3 \scriptsize{±0.4}\\
Soup-4   & 83.9 \scriptsize{±0.3}    & \enskip 1.9 \scriptsize{±0.1}       & 86.4 \scriptsize{±0.2}    & \enskip 1.4 \scriptsize{±0.1} & 88.6 \scriptsize{±0.3}    & \enskip 2.2 \scriptsize{±0.3}\\
Soup-8   & 84.0 \scriptsize{±0.2}    & \enskip 1.8 \scriptsize{±0.1}       & 86.4 \scriptsize{±0.2}    & \enskip 1.3 \scriptsize{±0.2} & 88.8 \scriptsize{±0.2}    & \enskip 2.2 \scriptsize{±0.3}\\
Soup-16   & 84.1 \scriptsize{±0.2}    & \enskip 1.7 \scriptsize{±0.1}       & 86.3 \scriptsize{±0.2}    & \enskip 1.3 \scriptsize{±0.1} & 88.9 \scriptsize{±0.2}    & \enskip 2.1 \scriptsize{±0.2}\\
\bottomrule
\end{tabular}
}
\caption{Results for the Add\_Data scenario on the test set. \textit{Ensemble-M} is the output ensemble of \textit{M} new models formed by averaging the probabilities. \textit{Soup-M} is the soup ensemble of \textit{M} new models formed by averaging the model weights.}
\label{tab:results_soup_add_data}
\end{table}
\begin{table}[t]
\centering
\resizebox{.999\linewidth}{!}{
\begin{tabular}{@{}lc@{}cc@{}cc@{}c@{}}
\toprule
& \multicolumn{2}{c}{\textbf{MASSIVE}}  & \multicolumn{2}{c}{\textbf{Banking77}} & \multicolumn{2}{c}{\textbf{AG News}}\\ 
\midrule
\textbf{Model} & \textbf{ACC}\textuparrow &  \textbf{NFR}\textdownarrow  & \textbf{ACC}\textuparrow & \textbf{NFR}\textdownarrow  & \textbf{ACC}\textuparrow & \textbf{NFR}\textdownarrow\\ 
\midrule
Old Model     & 68.8 \scriptsize{±0.1}    & \enskip 0.0 \scriptsize{±0.0}         & 80.0 \scriptsize{±0.3}    & \enskip 0.0 \scriptsize{±0.0} & 70.4 \scriptsize{±0.1}    & \enskip 0.0 \scriptsize{±0.0}\\
\midrule
Target Model   & 83.9 \scriptsize{±0.2}    & \enskip 3.2 \scriptsize{±0.2}       & 86.5 \scriptsize{±0.3}    & \enskip 2.8 \scriptsize{±0.3} & 87.9 \scriptsize{±0.5}    & \enskip 3.5 \scriptsize{±0.6}\\
New Model   & 83.8 \scriptsize{±0.2}    & \enskip 2.4 \scriptsize{±0.1}       & 86.3 \scriptsize{±0.3}    & \enskip 1.8 \scriptsize{±0.2} & 87.9 \scriptsize{±0.3}    & \enskip 4.2 \scriptsize{±0.5}\\
\midrule
Ensemble-2   & 83.9 \scriptsize{±0.2}    & \enskip 2.2 \scriptsize{±0.1}       & 86.8 \scriptsize{±0.3}    & \enskip 1.4 \scriptsize{±0.2} & 88.0 \scriptsize{±0.3}    & \enskip 4.2 \scriptsize{±0.4}\\
Ensemble-4   & 84.2 \scriptsize{±0.2}    & \enskip 2.0 \scriptsize{±0.1}       & 86.9 \scriptsize{±0.2}    & \enskip 1.2 \scriptsize{±0.1} & 88.0 \scriptsize{±0.3}    & \enskip 4.3 \scriptsize{±0.4}\\
Ensemble-8   & 84.2 \scriptsize{±0.2}    & \enskip 1.9 \scriptsize{±0.2}       & 87.0 \scriptsize{±0.3}    & \enskip 1.1 \scriptsize{±0.1} & 88.0 \scriptsize{±0.3}    & \enskip 4.2 \scriptsize{±0.3}\\
Ensemble-16   & 84.3 \scriptsize{±0.2}    & \enskip 1.9 \scriptsize{±0.2}       & 87.1 \scriptsize{±0.3}    & \enskip 1.0 \scriptsize{±0.1} & 88.1 \scriptsize{±0.3}    & \enskip 4.2 \scriptsize{±0.3}\\
\midrule
Soup-2   & 83.9 \scriptsize{±0.3}    & \enskip 2.1 \scriptsize{±0.1}       & 86.7 \scriptsize{±0.3}    & \enskip 1.4 \scriptsize{±0.2} & 88.0 \scriptsize{±0.3}    & \enskip 4.2 \scriptsize{±0.3}\\
Soup-4   & 84.0 \scriptsize{±0.2}    & \enskip 1.9 \scriptsize{±0.1}       & 86.8 \scriptsize{±0.3}    & \enskip 1.1 \scriptsize{±0.1} & 88.0 \scriptsize{±0.3}    & \enskip 4.2 \scriptsize{±0.4}\\
Soup-8   & 84.1 \scriptsize{±0.2}    & \enskip 1.8 \scriptsize{±0.2}       & 86.9 \scriptsize{±0.2}    & \enskip 1.0 \scriptsize{±0.1} & 88.1 \scriptsize{±0.3}    & \enskip 4.2 \scriptsize{±0.4}\\
Soup-16   & 84.1 \scriptsize{±0.2}    & \enskip 1.8 \scriptsize{±0.2}       & 86.9 \scriptsize{±0.2}    & \enskip 1.0 \scriptsize{±0.1} & 88.1 \scriptsize{±0.3}    & \enskip 4.1 \scriptsize{±0.4}\\
\bottomrule
\end{tabular}
}
\caption{Results for the Add\_Classes scenario on the test set of the three datasets. \textit{Ensemble-M} is the output ensemble of \textit{M} new models formed by averaging the probabilities. \textit{Soup-M} is the soup ensemble of \textit{M} new models formed by averaging the model weights.}
\label{tab:results_soup_add_classes}
\end{table}

\begin{figure}[ht]
    \centering
    \includegraphics[width=0.99\linewidth]{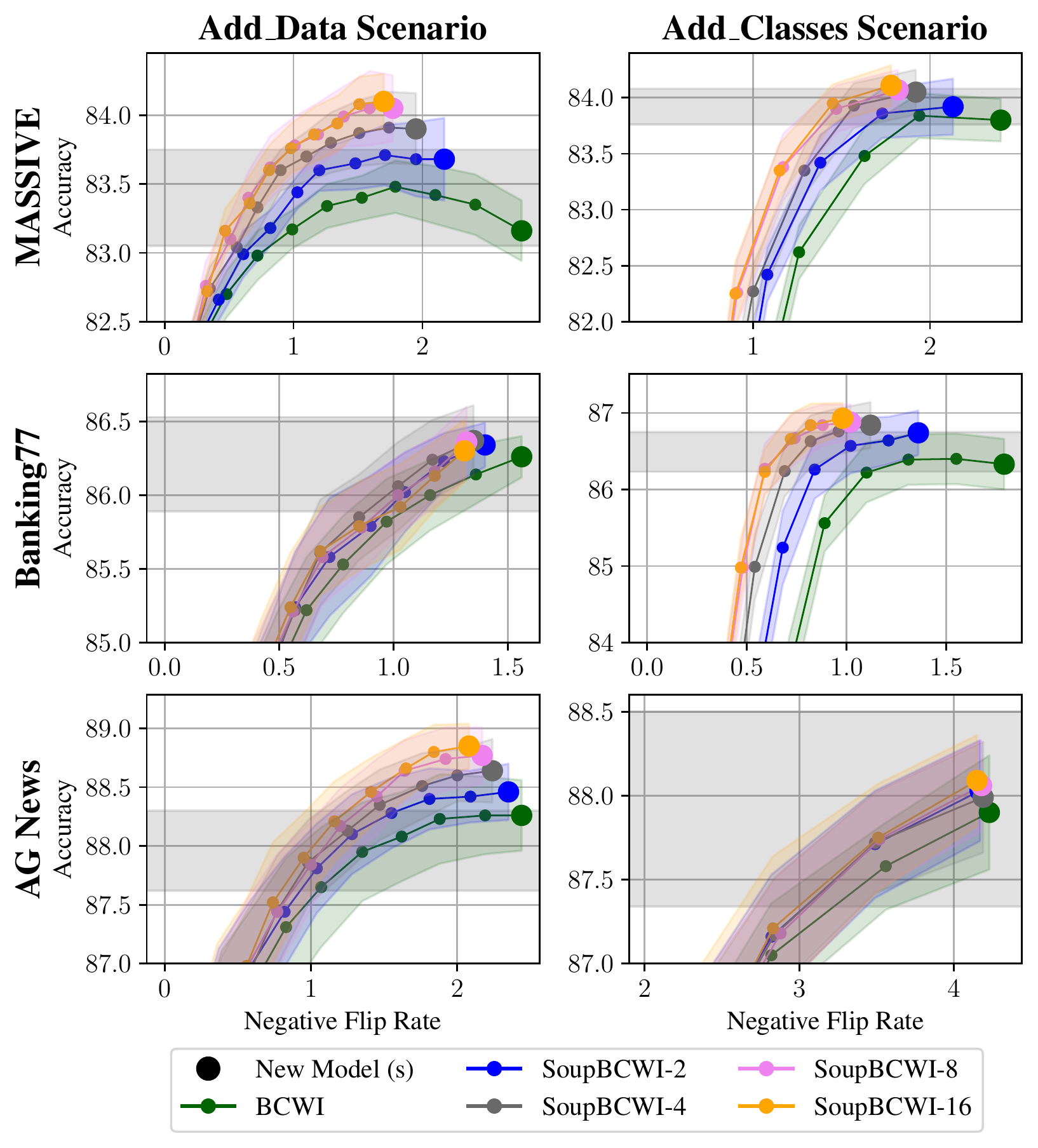}
    \caption{Plots for SoupBCWI in comparison with vanilla BCWI. The gray area indicates the target accuracy level.}
    \label{fig:SoupBCWI}
\end{figure}

\section{Access to Old Data}
\begin{figure*}[ht]
    \centering
    \includegraphics[width=0.95\textwidth]{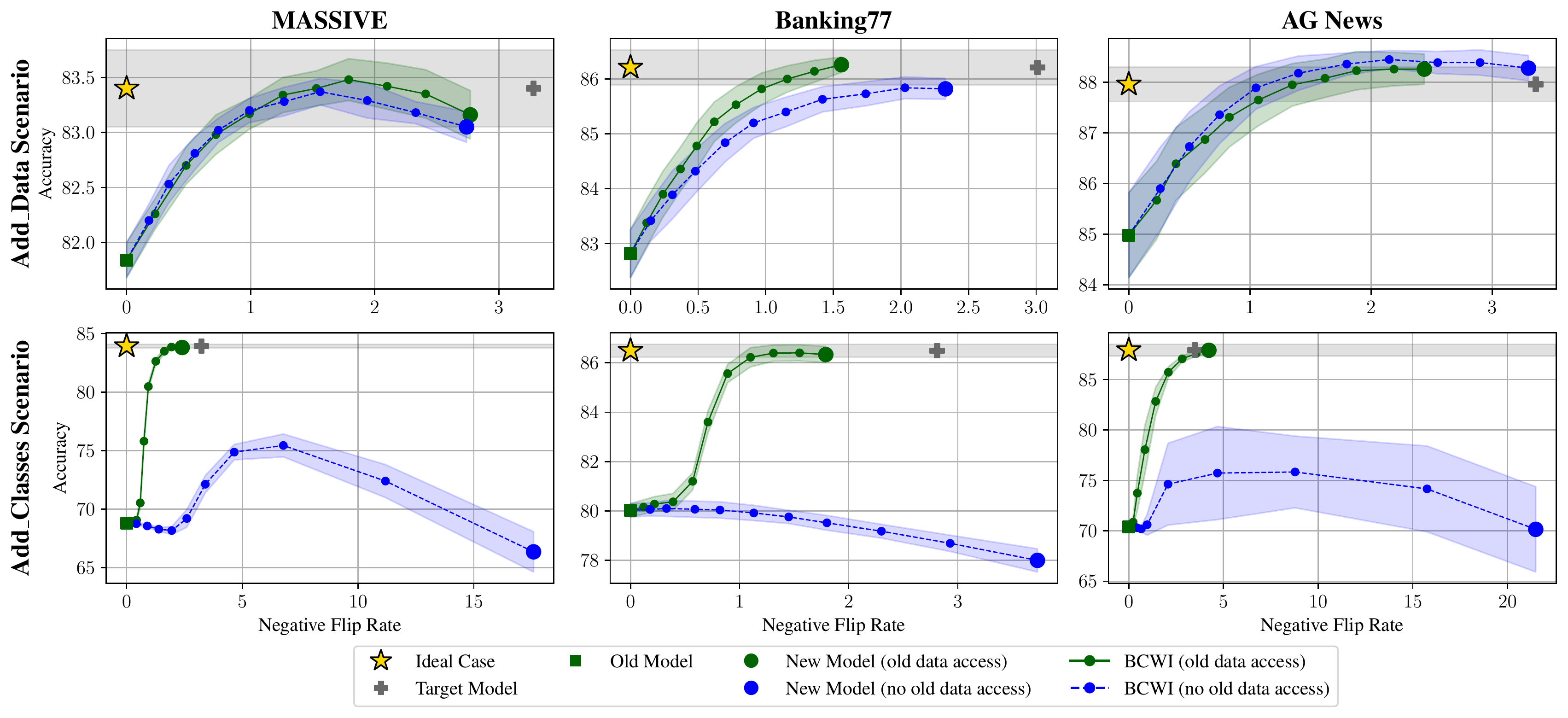}
    \caption{Comparison of BCWI with and without access to the old data during training of the new model. Results for the AD and AC scenario evaluated on our test sets of MASSIVE~\cite{FitzGerald2022MASSIVEA1}, Banking77~\cite{Casanueva2020EfficientID} and AG~News~\cite{Zhang2015CharacterlevelCN}. The gray horizontal bar is spanned by the 95\% confidence interval of the target model and indicates the the level of accuracy a model should reach. The $\alpha$-values are in 0.1 steps where 0.0 is equivalent to the new model and 1.0 is equivalent to the old model. The ideal case for a model is to have zero negative flips while maintaining the target accuracy.}
    \label{fig:no_access_plots}
\end{figure*}
For our main experiments we assume full access to the old data. This allows us to train the new model without catastrophic forgetting. To complement these results, we also show the behavior of BCWI when the new model is trained only on the new data (i.e.~no access to the old data). The results are presented in Figure~\ref{fig:no_access_plots} and show that for the AD scenario the more restrictive setting has negative impact on Banking77 but achieves similar results for MASSIVE and AG~News. In the the AC scenario, the new model has significantly lower accuracy which can be attributed to catastrophic forgetting, because the new model is finetuned on new classes only. The interpolation towards the old model improves accuracy but does not reach the same accuracy as finetuning the new model on old and new data.

\section{Datasets and Scenarios}
\begin{figure*}
  \centering
  \begin{subfigure}{0.3\textwidth}
  \includegraphics[width=\textwidth]{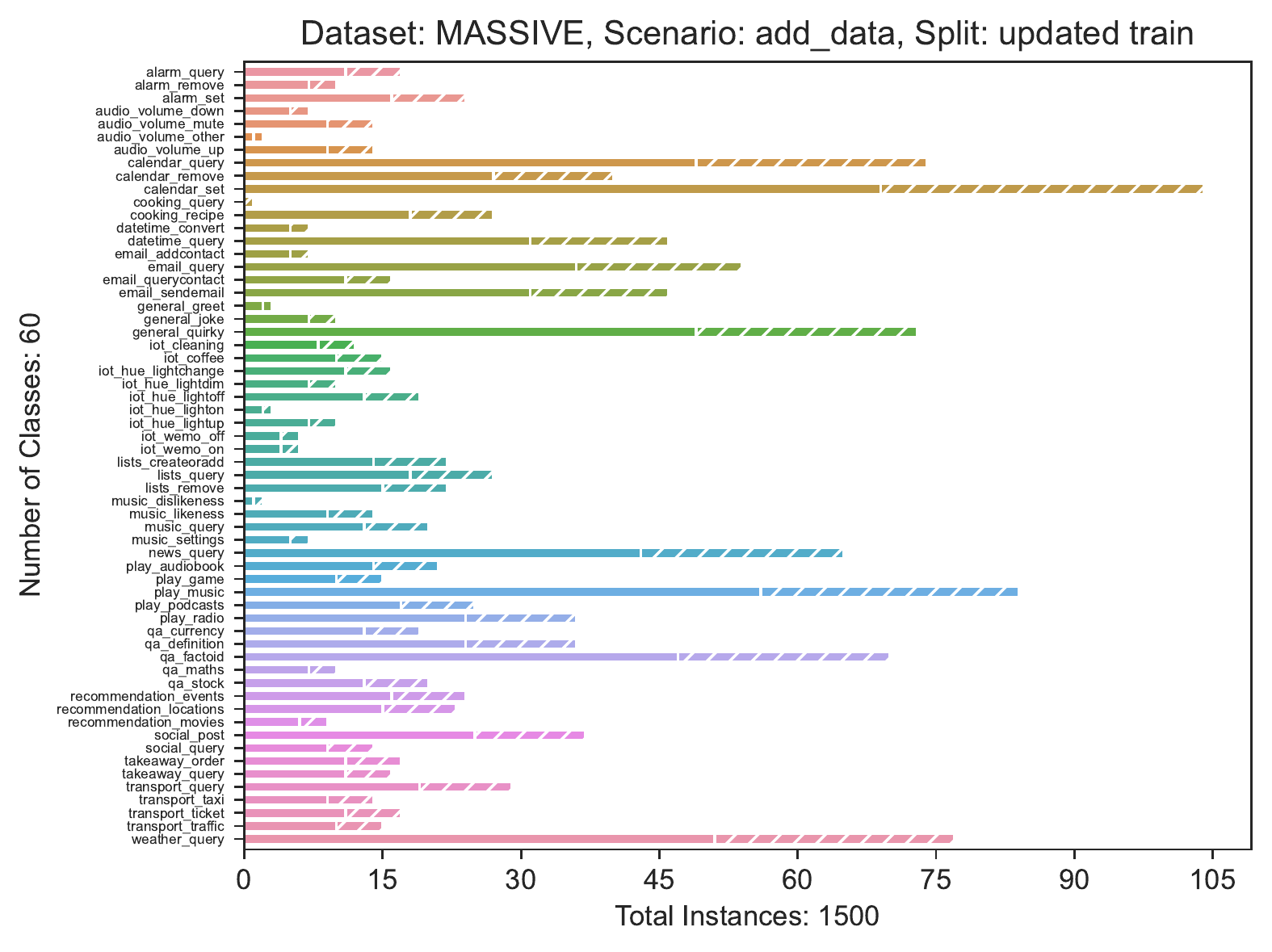}
  \end{subfigure}
  \hfill
  \begin{subfigure}{0.3\textwidth}
  \includegraphics[width=\textwidth]{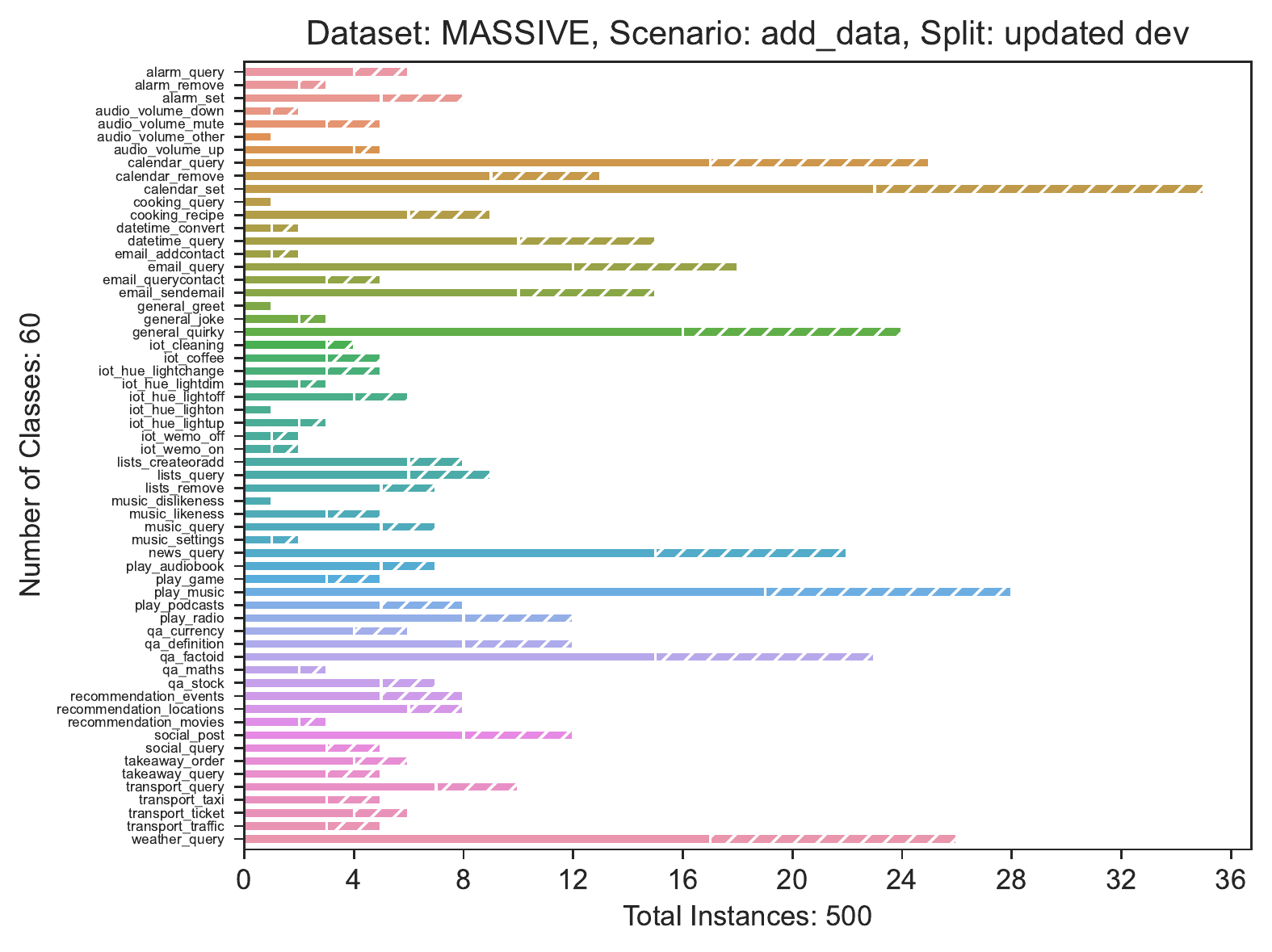}
  \end{subfigure}
  \hfill
  \begin{subfigure}{0.3\textwidth}
  \includegraphics[width=\textwidth]{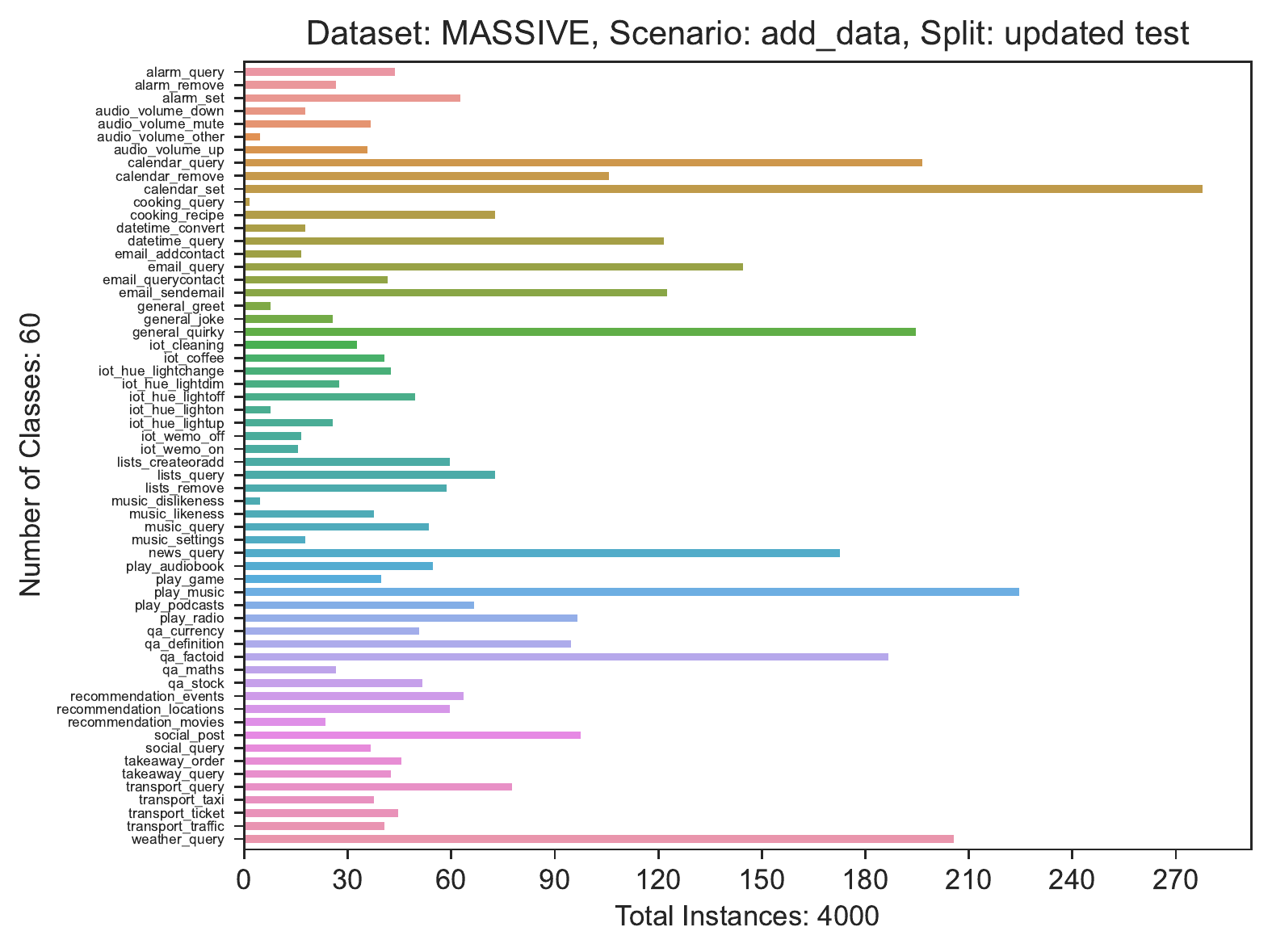}
  \end{subfigure}
  \begin{subfigure}{0.3\textwidth}
  \includegraphics[width=\textwidth]{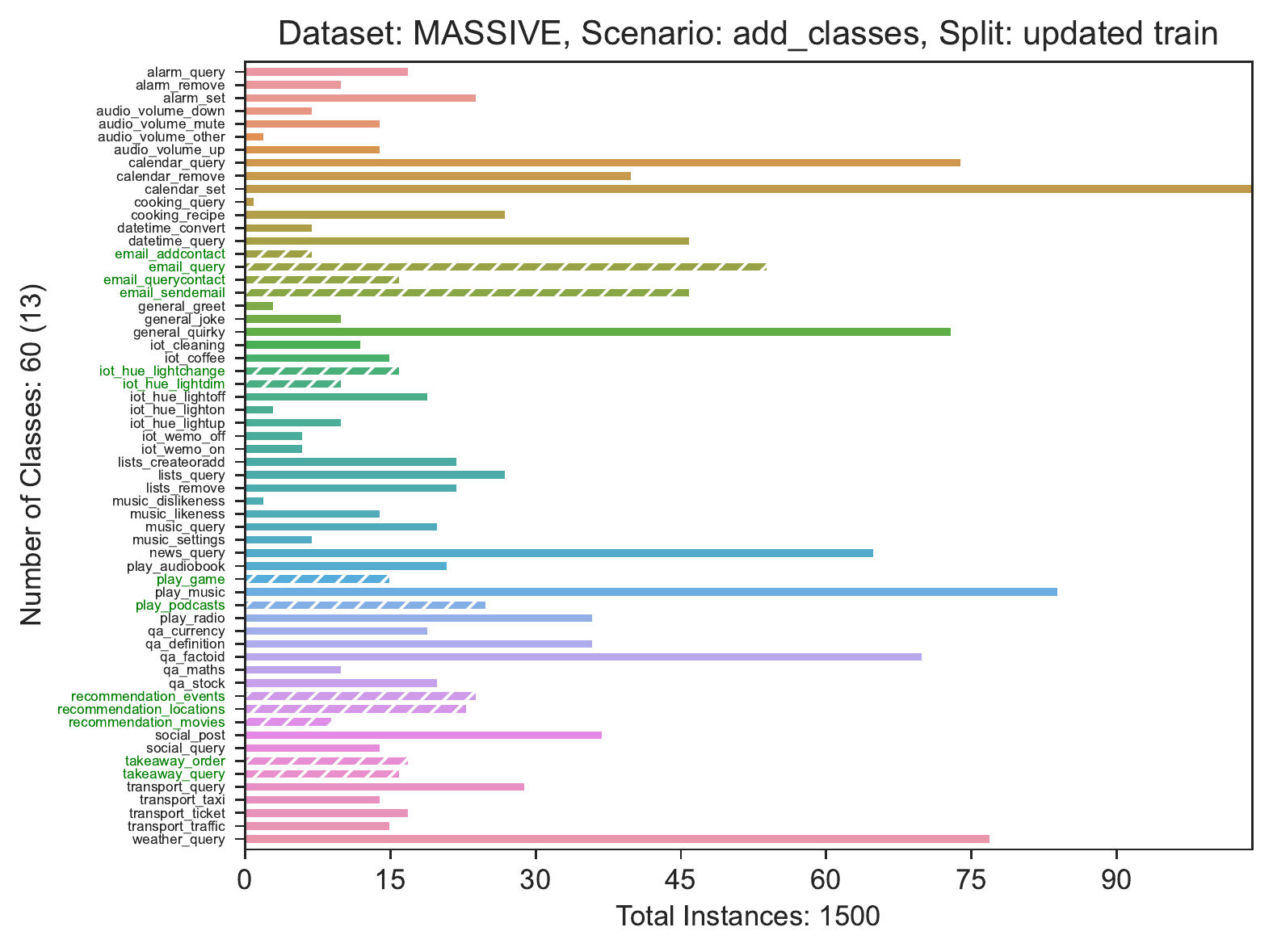}
  \end{subfigure}
  \hfill
  \begin{subfigure}{0.3\textwidth}
  \includegraphics[width=\textwidth]{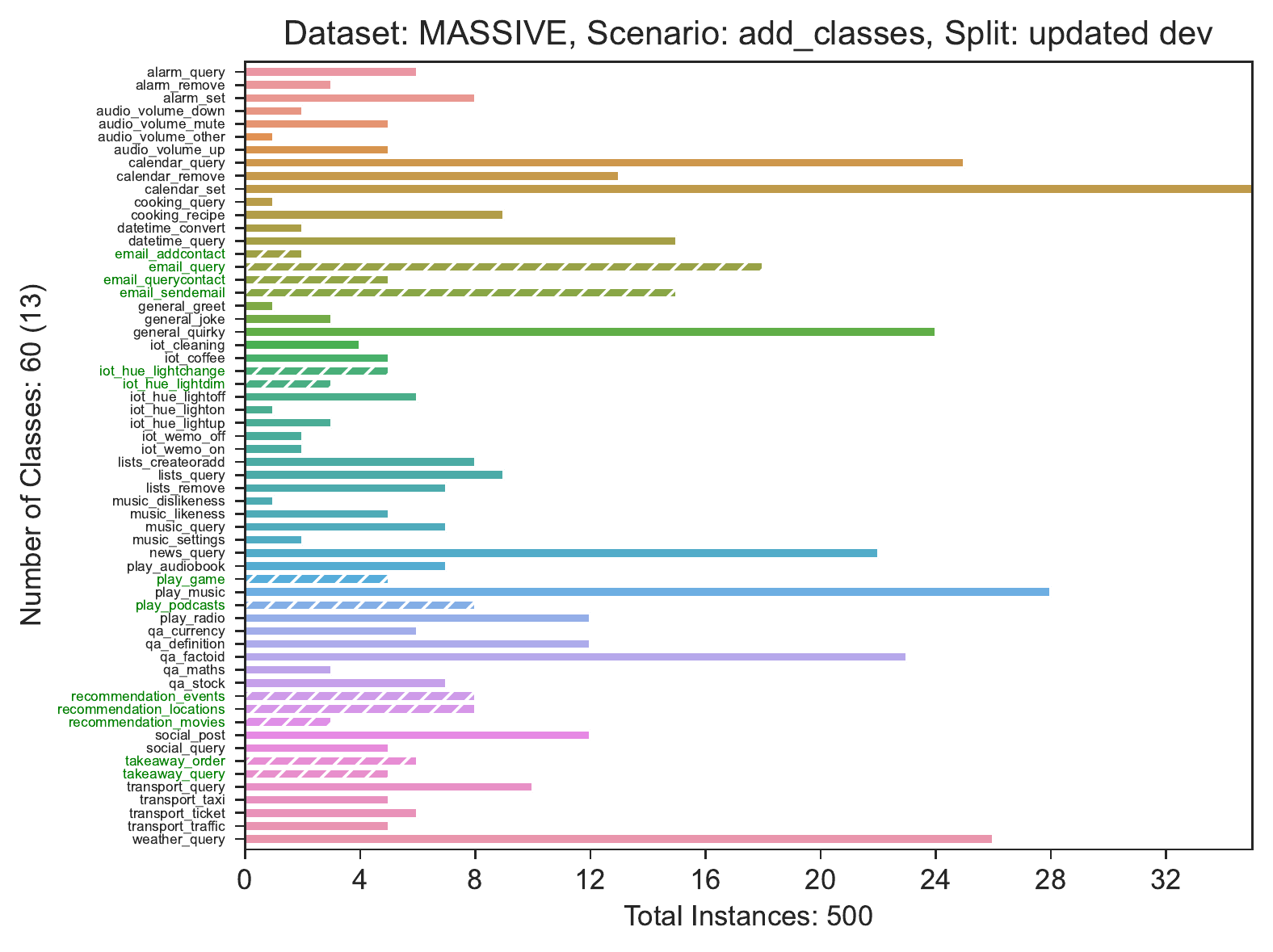}
  \end{subfigure}
  \hfill
  \begin{subfigure}{0.3\textwidth}
  \includegraphics[width=\textwidth]{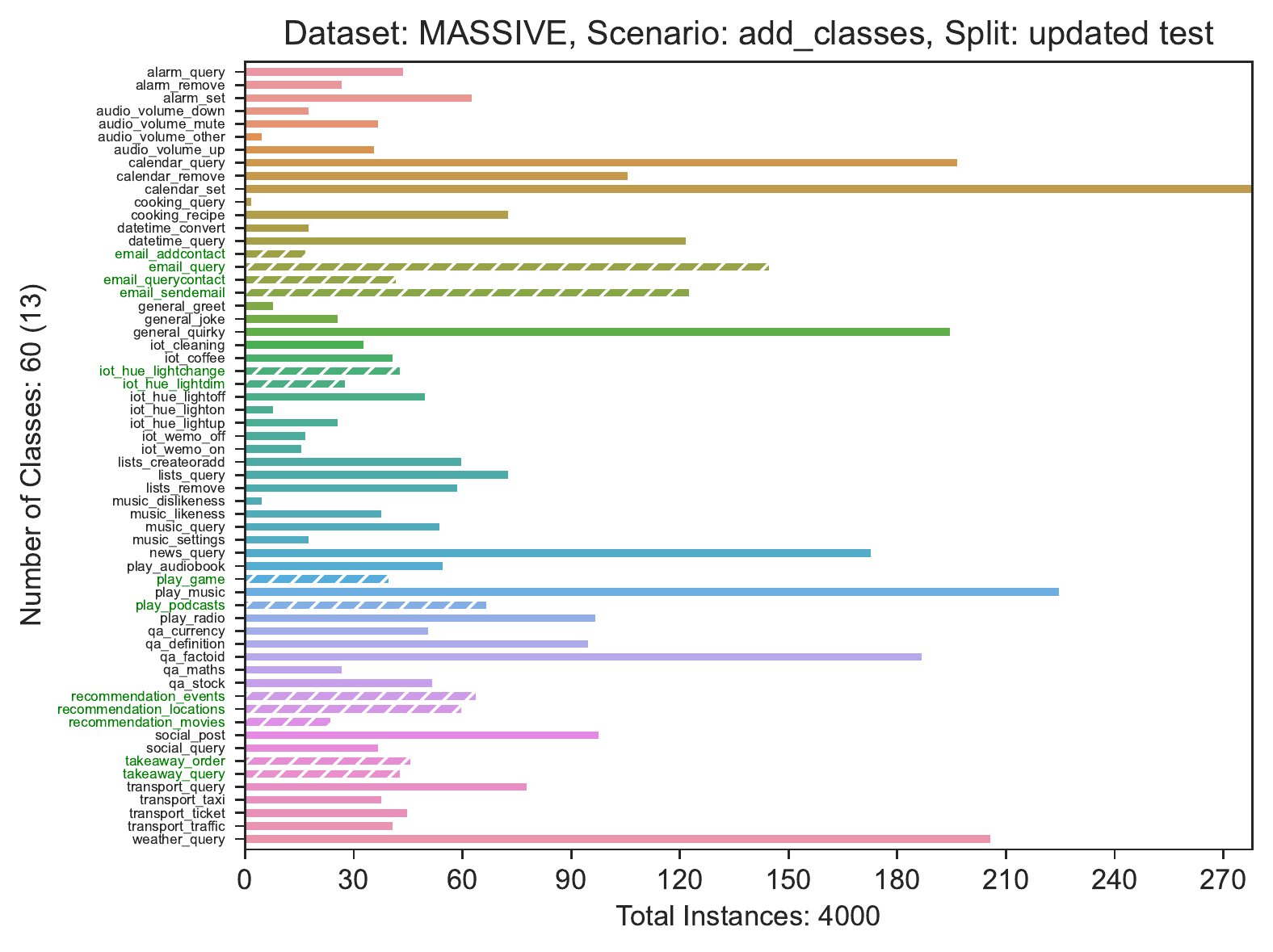}
  \end{subfigure}
  
  \begin{subfigure}{0.3\textwidth}
  \includegraphics[width=\textwidth]{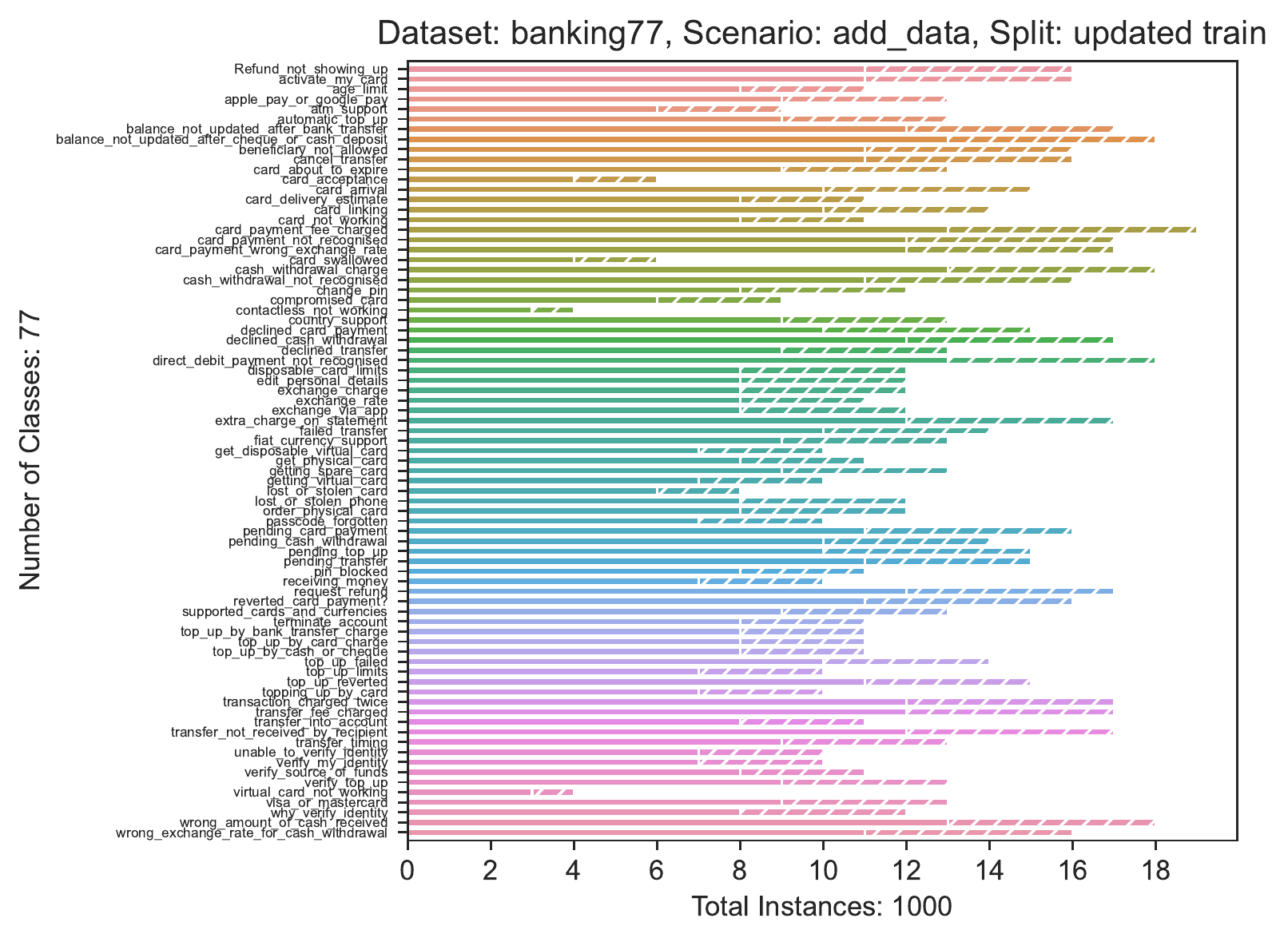}
  \end{subfigure}
  \hfill
  \begin{subfigure}{0.3\textwidth}
  \includegraphics[width=\textwidth]{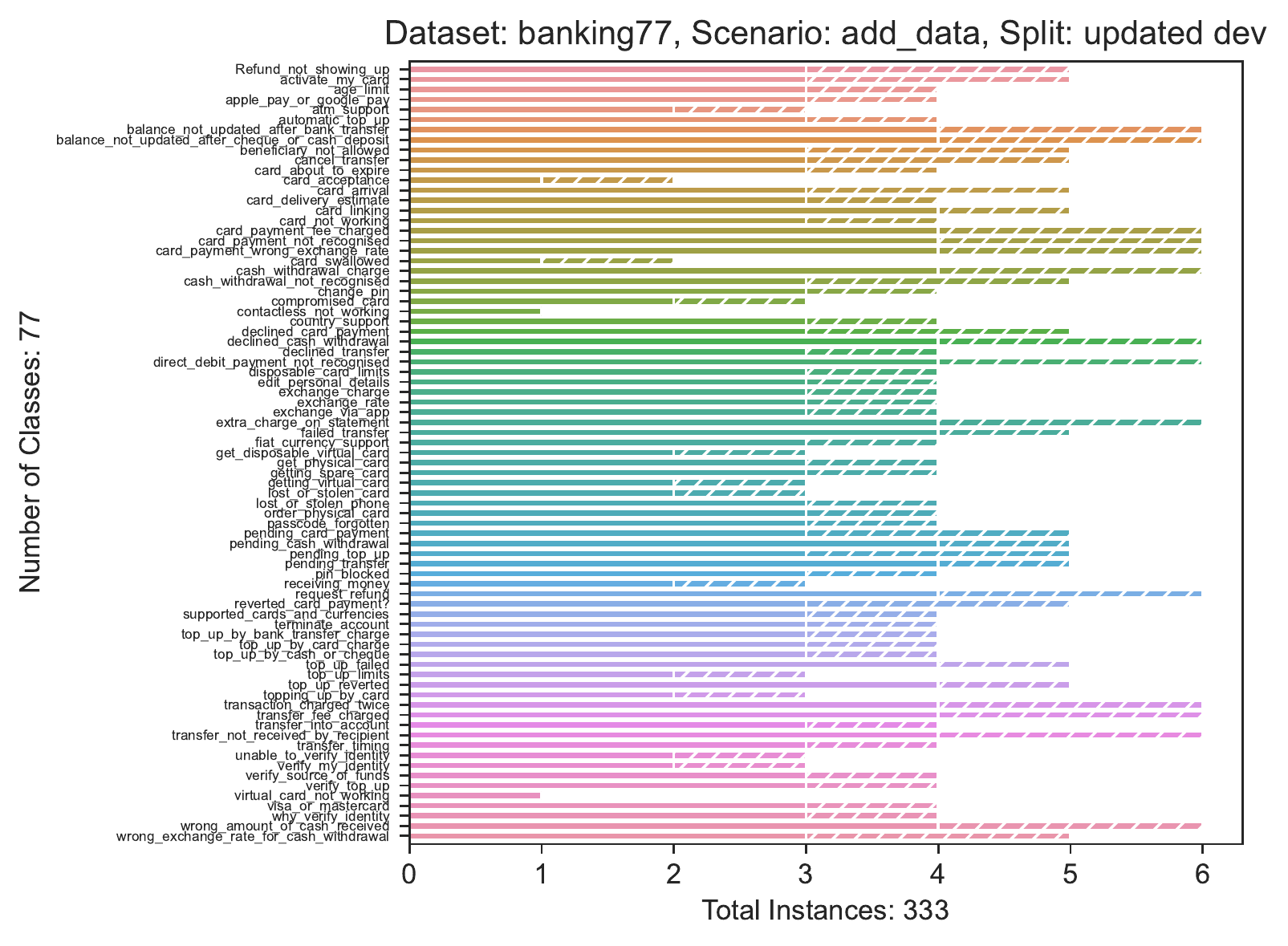}
  \end{subfigure}
  \hfill
  \begin{subfigure}{0.3\textwidth}
  \includegraphics[width=\textwidth]{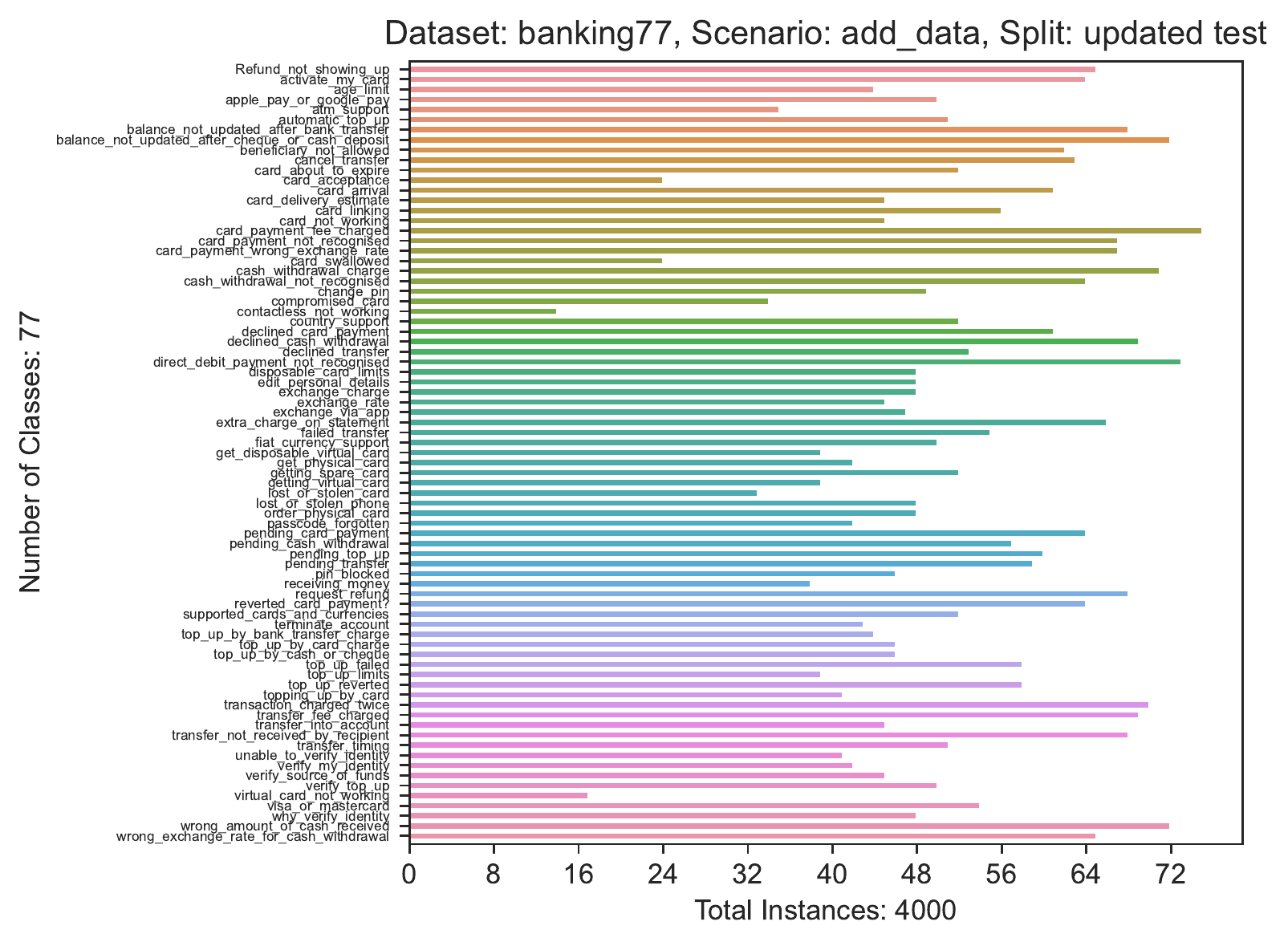}
  \end{subfigure}
  \begin{subfigure}{0.3\textwidth}
  \includegraphics[width=\textwidth]{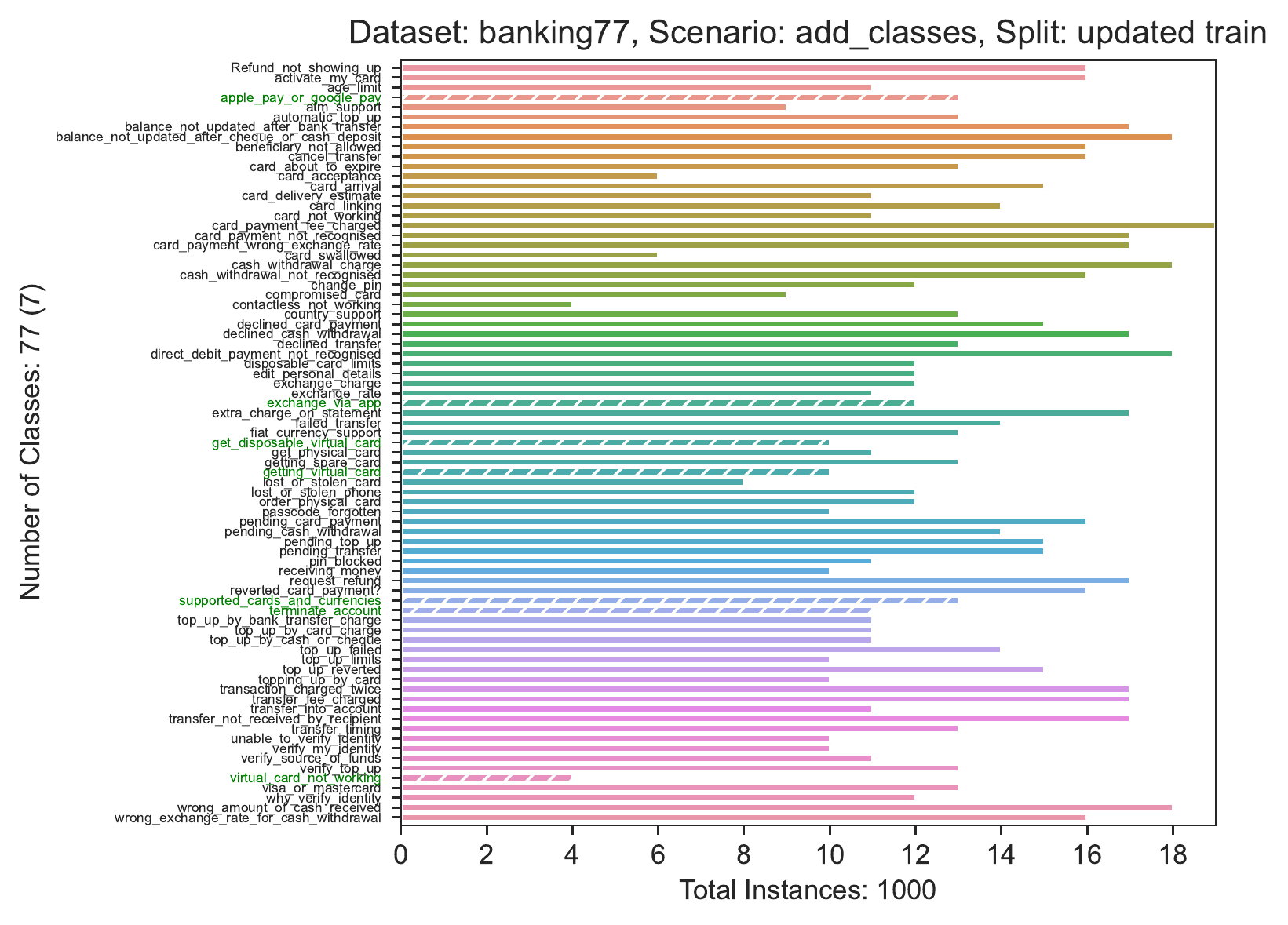}
  \end{subfigure}
  \hfill
  \begin{subfigure}{0.3\textwidth}
  \includegraphics[width=\textwidth]{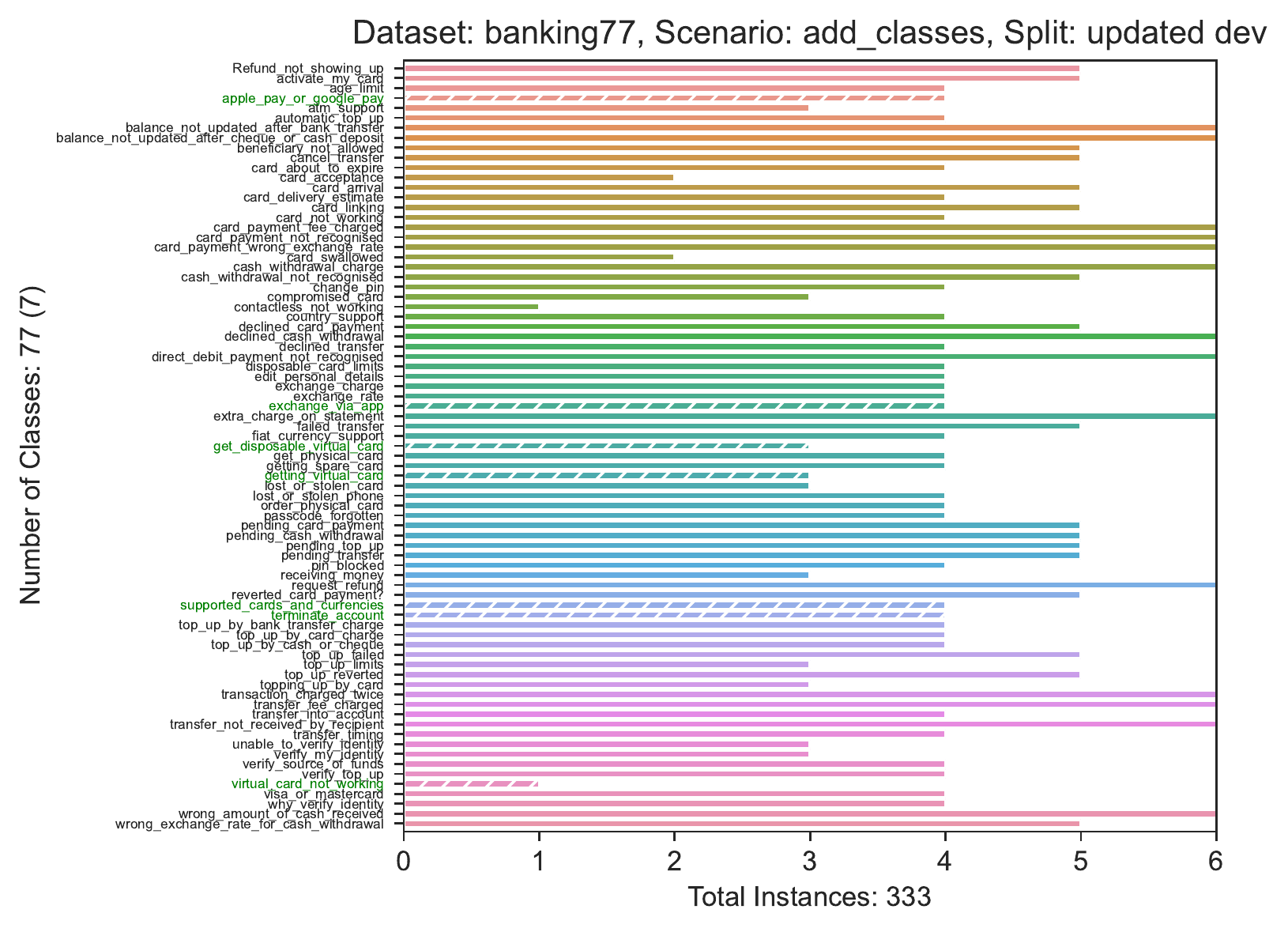}
  \end{subfigure}
  \hfill
  \begin{subfigure}{0.3\textwidth}
  \includegraphics[width=\textwidth]{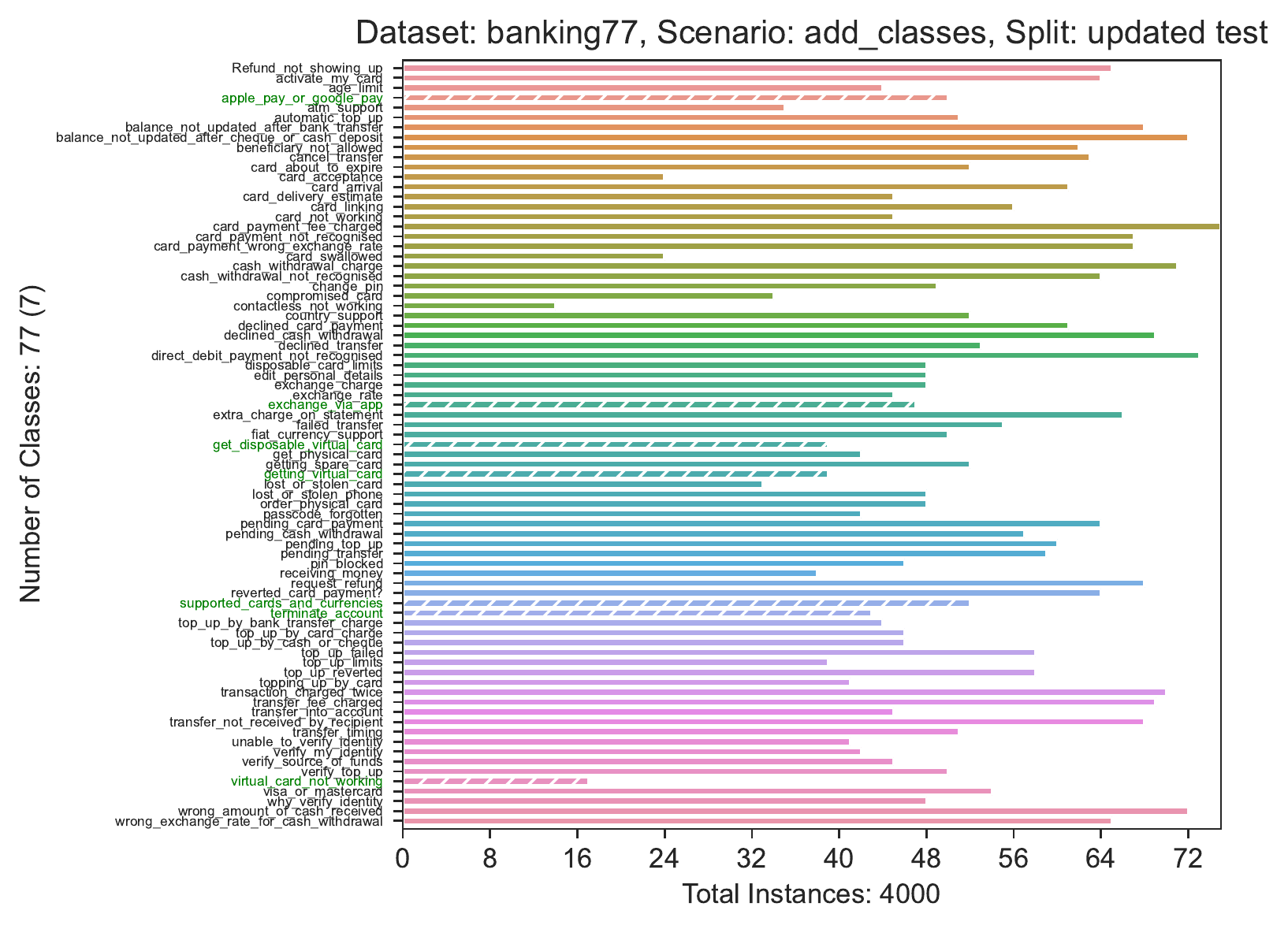}
  \end{subfigure}
  
  \begin{subfigure}{0.3\textwidth}
  \includegraphics[width=\textwidth]{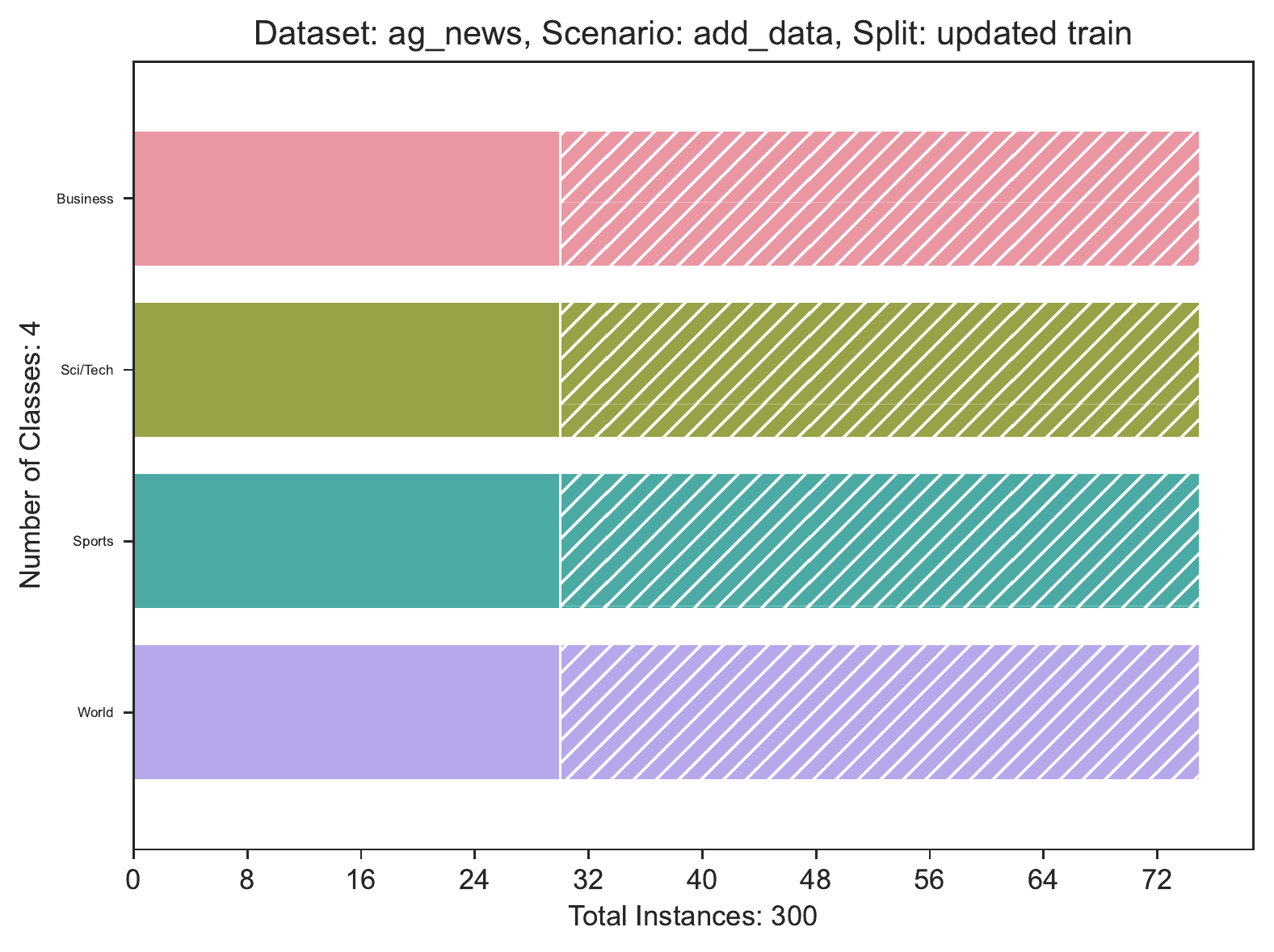}
  \end{subfigure}
  \hfill
  \begin{subfigure}{0.3\textwidth}
  \includegraphics[width=\textwidth]{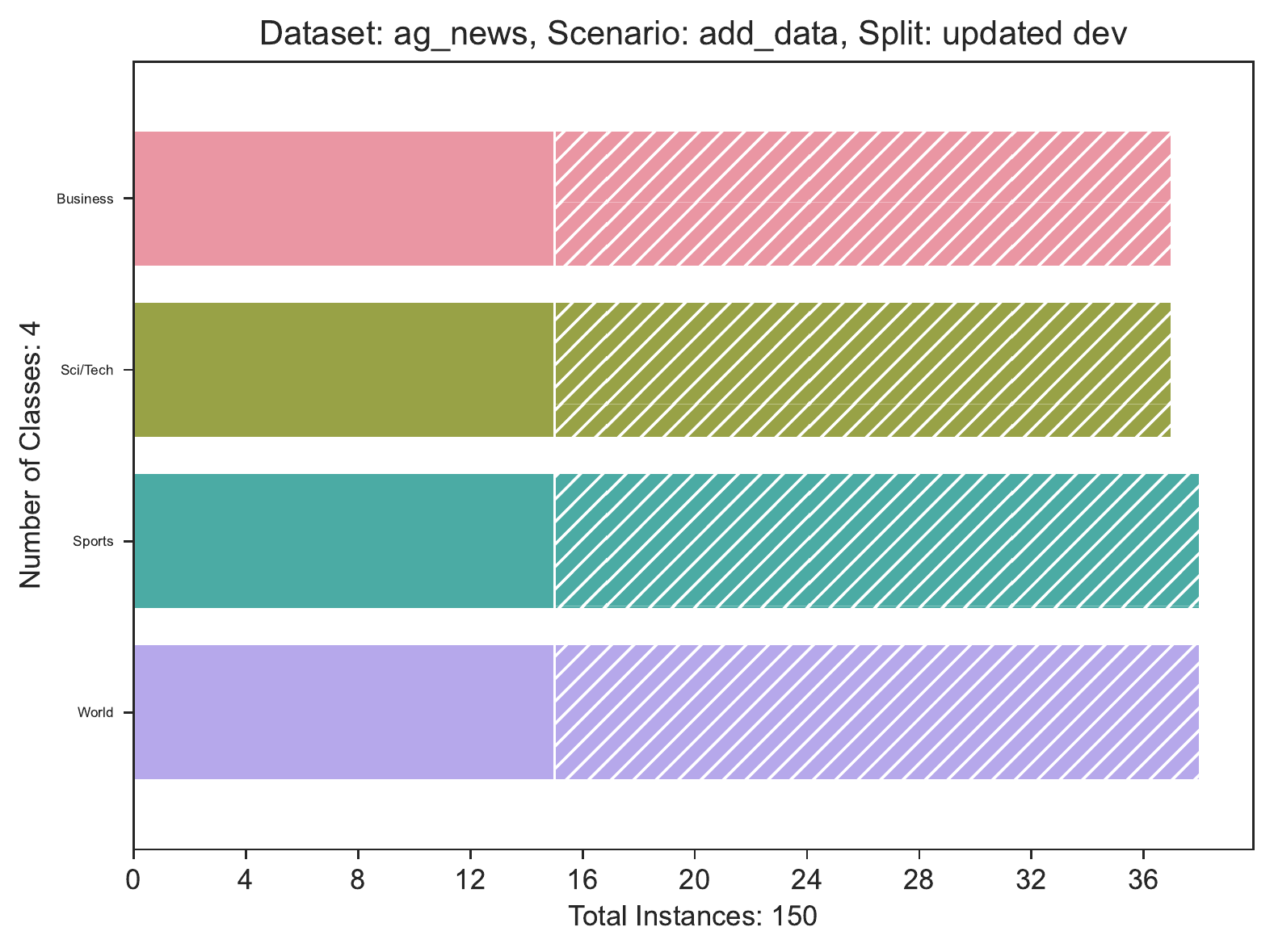}
  \end{subfigure}
  \hfill
  \begin{subfigure}{0.3\textwidth}
  \includegraphics[width=\textwidth]{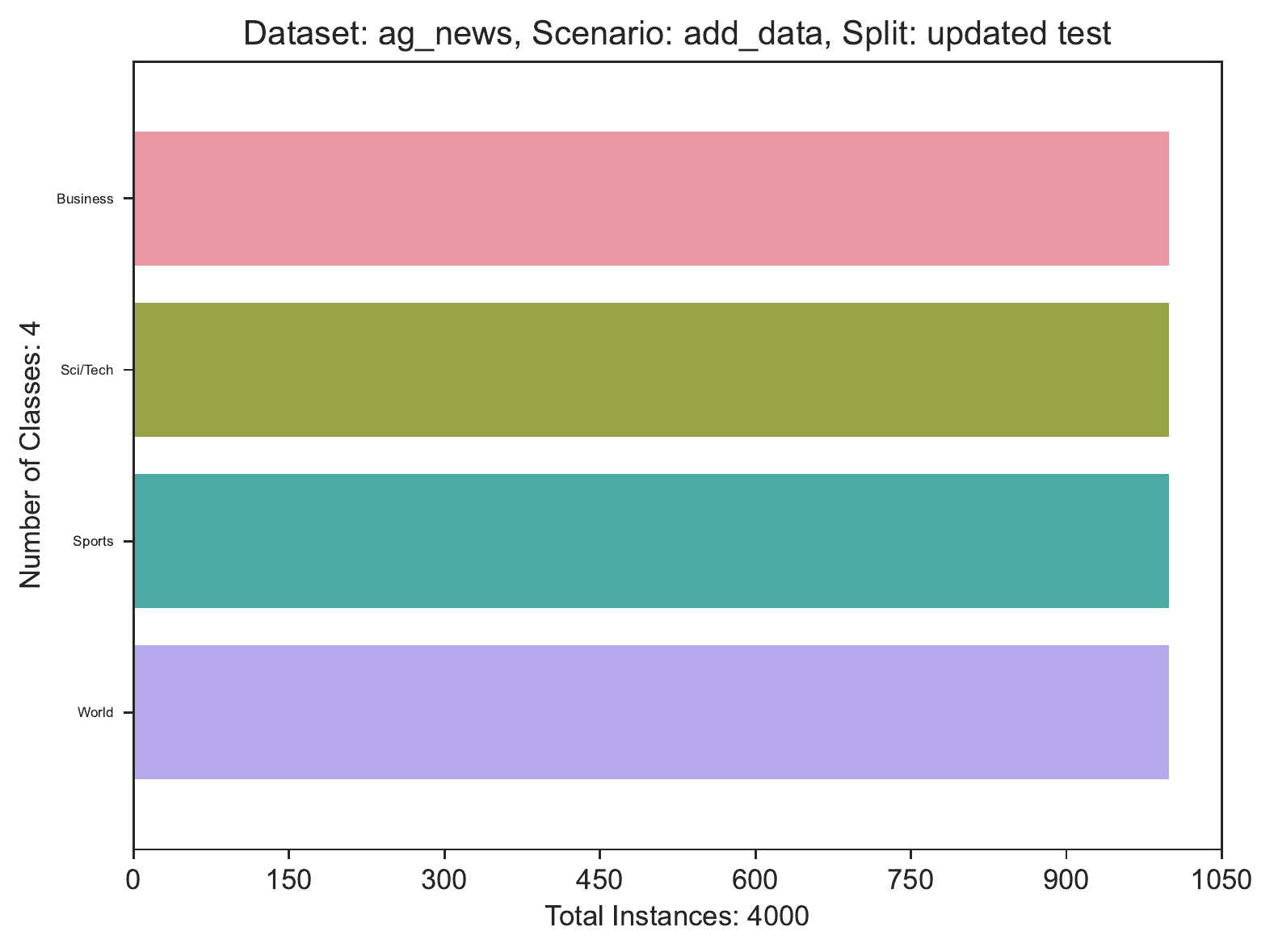}
  \end{subfigure}
  \begin{subfigure}{0.3\textwidth}
  \includegraphics[width=\textwidth]{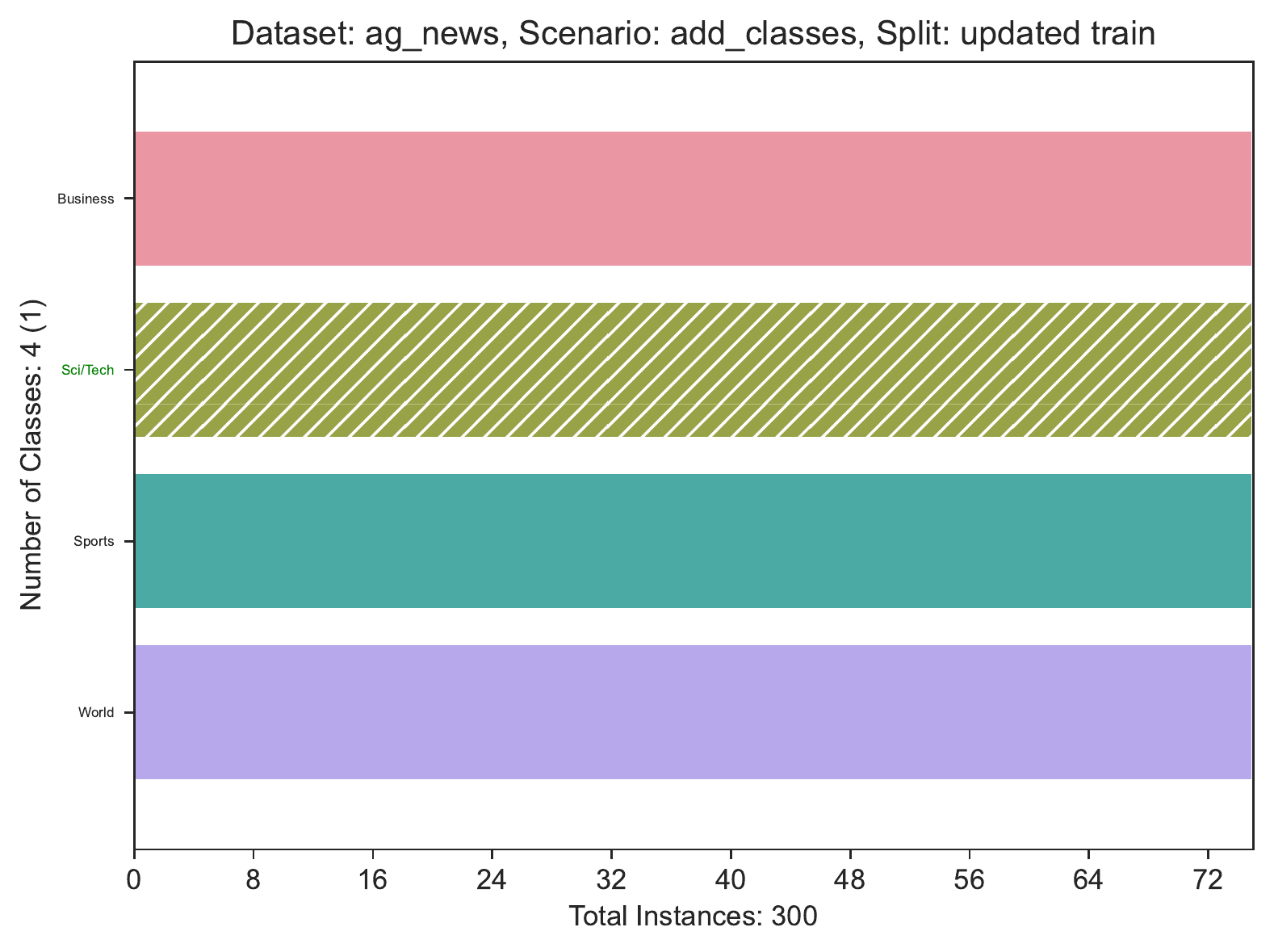}
  \end{subfigure}
  \hfill
  \begin{subfigure}{0.3\textwidth}
  \includegraphics[width=\textwidth]{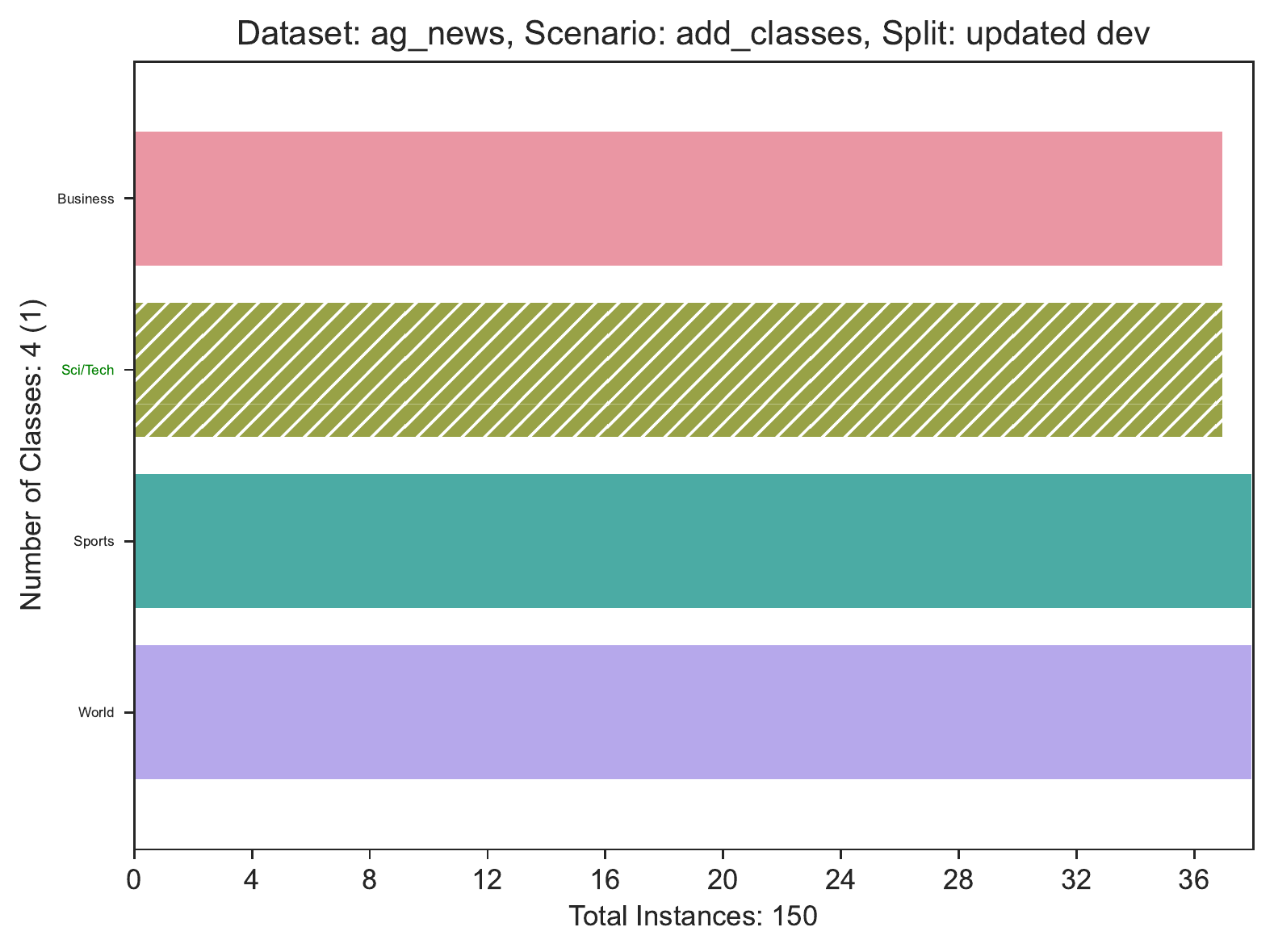}
  \end{subfigure}
  \hfill
  \begin{subfigure}{0.3\textwidth}
  \includegraphics[width=\textwidth]{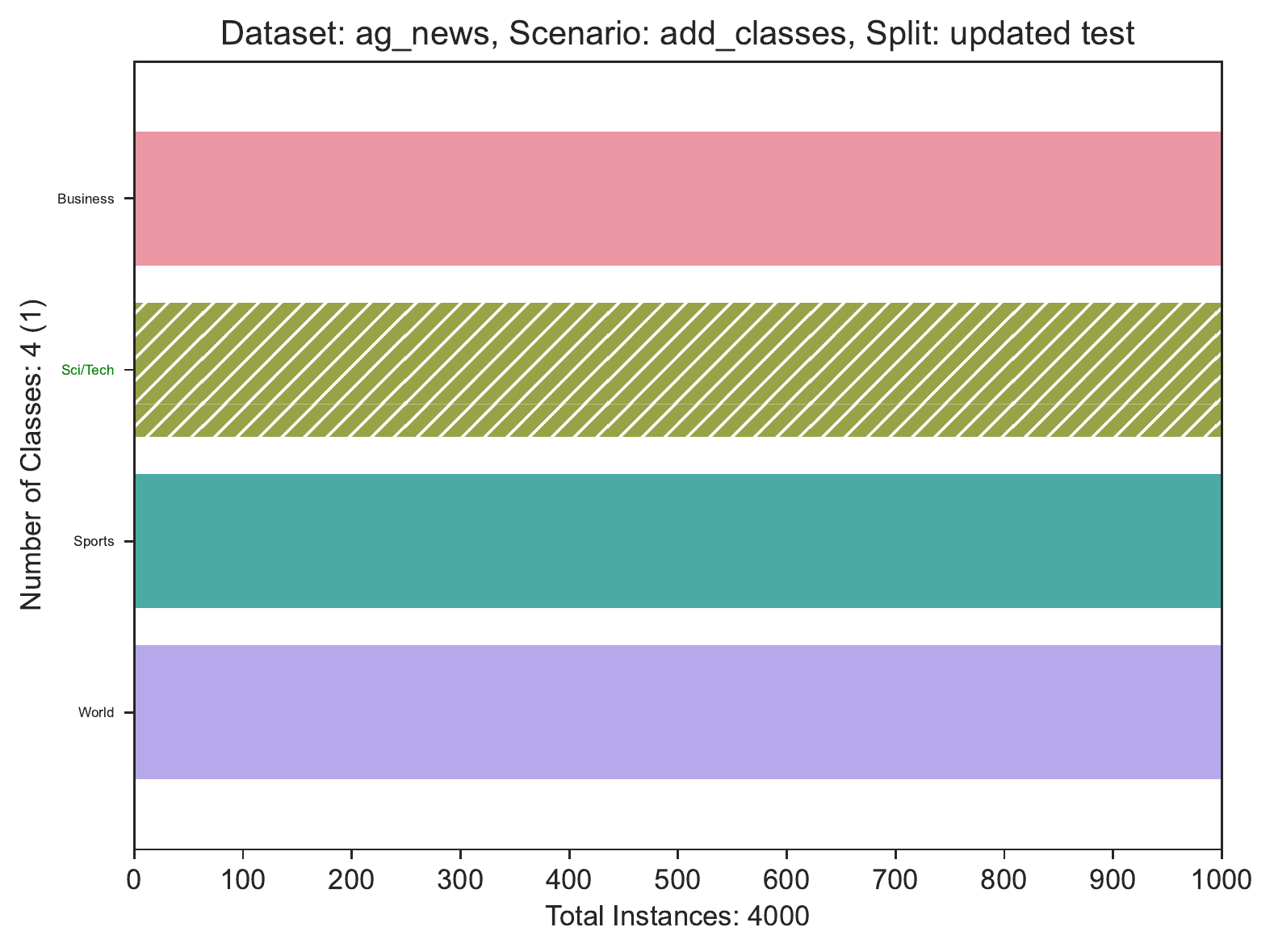}
  \end{subfigure}
  \caption{Add\_Data scenario and Add\_Classes scenario for MASSIVE~\cite{FitzGerald2022MASSIVEA1}, Banking77~\cite{Casanueva2020EfficientID} and AG~News~\cite{Zhang2015CharacterlevelCN}. Striped bars indicate added instances. Added class names are printed in green.}
  \label{fig:data_distributions}
\end{figure*}

Detailed label distribution and number of instances for the AD and AC scenarios for all three datasets are visualized in Figure~\ref{fig:data_distributions}. The plots also show which classes are added for each dataset in the AC scenario.

\clearpage
\begin{table*}[t]
\centering
\resizebox{.999\linewidth}{!}{
\begin{tabular}{@{}l|cc@{}cc@{}c|cc@{}cc@{}c|cc@{}cc@{}c@{}}
\toprule
& \multicolumn{5}{c|}{\textbf{MASSIVE}}  & \multicolumn{5}{c|}{\textbf{Banking77}} & \multicolumn{5}{c}{\textbf{AG News}}\\ 
\midrule
 & $\lambda$ & \multicolumn{2}{c}{\textbf{dev}} &  \multicolumn{2}{c|}{\textbf{test}}  & $\lambda$ &  \multicolumn{2}{c}{\textbf{dev}} & \multicolumn{2}{c|}{\textbf{test}}  & $\lambda$ &  \multicolumn{2}{c}{\textbf{dev}} & \multicolumn{2}{c}{\textbf{test}} \\
\midrule
\textbf{Model} &  & \textbf{ACC}\textuparrow &  \textbf{NFR}\textdownarrow  & \textbf{ACC}\textuparrow & \textbf{NFR}\textdownarrow  &  &  \textbf{ACC}\textuparrow & \textbf{NFR}\textdownarrow & \textbf{ACC}\textuparrow &  \textbf{NFR}\textdownarrow  &  &  \textbf{ACC}\textuparrow & \textbf{NFR}\textdownarrow  & \textbf{ACC}\textuparrow & \textbf{NFR}\textdownarrow\\
\midrule
Old Model     & - & 80.4 \scriptsize{±0.8}    & \enskip 0.0 \scriptsize{±0.0}         & 81.8 \scriptsize{±0.2}    & \enskip 0.0 \scriptsize{±0.0} & - &  83.2 \scriptsize{±0.9}    & \enskip 0.0 \scriptsize{±0.0} & 82.8 \scriptsize{±0.4}    & \enskip 0.0 \scriptsize{±0.0}         & - &  84.1 \scriptsize{±1.5}    & \enskip 0.0 \scriptsize{±0.0} & 85.0 \scriptsize{±0.8}    & \enskip 0.0 \scriptsize{±0.0}\\
Target Model     & - & 82.2 \scriptsize{±0.4}    & \enskip 3.0 \scriptsize{±0.6}         & 83.4 \scriptsize{±0.4}    & \enskip 3.3 \scriptsize{±0.4} & - &  86.4 \scriptsize{±0.7}    & \enskip 2.8 \scriptsize{±0.8} & 86.2 \scriptsize{±0.4}    & \enskip 3.0 \scriptsize{±0.3}         & - &  88.5 \scriptsize{±1.1}    & \enskip 3.1 \scriptsize{±0.9} & 88.0 \scriptsize{±0.1}    & \enskip 3.4 \scriptsize{±0.3}\\
New Model     & - & 82.0 \scriptsize{±1.0}    & \enskip 2.4 \scriptsize{±0.5}         & 83.2 \scriptsize{±0.2}    & \enskip 2.8 \scriptsize{±0.2} & - &  86.1 \scriptsize{±0.8}   & \enskip \textbf{1.1} \scriptsize{±0.3} & 86.3 \scriptsize{±0.1}    & \enskip 1.6 \scriptsize{±0.1}         & - &  89.5 \scriptsize{±0.8}    & \enskip 1.3 \scriptsize{±0.3} & 88.3 \scriptsize{±0.3}    & \enskip 2.4 \scriptsize{±0.3}\\
\midrule
ACC Threshold & \multicolumn{2}{c}{\enskip\smallbreak$\geq$ 81.8} & &&&\multicolumn{2}{c}{\enskip\enskip$\geq$ 85.8} & &&& \multicolumn{2}{c}{\enskip\enskip$\geq$ 89.0} & &&\\
\midrule
\midrule
PriorWD     & 100 & 81.8 \scriptsize{±0.7}    & \enskip \textbf{1.7} \scriptsize{±0.3}         & 83.4 \scriptsize{±0.3}    & \enskip 2.0 \scriptsize{±0.2} & 200 &  86.1 \scriptsize{±0.7}   & \enskip \textbf{0.8} \scriptsize{±0.3} & 85.9 \scriptsize{±0.3}    & \enskip 1.3 \scriptsize{±0.1}         & 1e3 &  89.5 \scriptsize{±0.8}    & \enskip \textbf{0.8} \scriptsize{±0.5} & 88.1 \scriptsize{±0.4}    & \enskip \textbf{1.7} \scriptsize{±0.2}\\
Mixout     & 0.2 & 81.8 \scriptsize{±0.5}    & \enskip 2.2 \scriptsize{±0.4}         & 83.0 \scriptsize{±0.2}    & \enskip 2.6 \scriptsize{±0.2} & 0.9 &  86.1 \scriptsize{±0.7}   & \enskip \textbf{0.9} \scriptsize{±0.2} & 85.8 \scriptsize{±0.3}    & \enskip 1.4 \scriptsize{±0.1}         & 0.95 &  89.7 \scriptsize{±0.9}    & \enskip \textbf{0.9} \scriptsize{±0.6} & 88.4 \scriptsize{±0.4}    & \enskip \textbf{1.6} \scriptsize{±0.2}\\
EWC     & 0.01 & 82.0 \scriptsize{±0.8}    & \enskip \textbf{1.8} \scriptsize{±0.4}         & 83.3 \scriptsize{±0.3}    & \enskip 2.1 \scriptsize{±0.2} & 0.01 &  86.4 \scriptsize{±1.0}   & \enskip \textbf{0.8} \scriptsize{±0.2} & 86.1 \scriptsize{±0.2}    & \enskip 1.4 \scriptsize{±0.1}         & 1.0 & 88.9 \scriptsize{±1.0}  & \enskip \textbf{0.9} \scriptsize{±0.6}          & 87.9 \scriptsize{±0.4}    & \enskip \textbf{1.6} \scriptsize{±0.3}\\
\midrule
BCWI     & 0.45 & 81.8 \scriptsize{±0.7}    & \enskip \textbf{1.2} \scriptsize{±0.3}         & 83.4 \scriptsize{±0.1}    & \enskip \textbf{1.4} \scriptsize{±0.1} & 0.4 &  85.8 \scriptsize{±0.8}   & \enskip \textbf{0.6} \scriptsize{±0.2} & 85.5 \scriptsize{±0.3}    & \enskip \textbf{0.8} \scriptsize{±0.1}         & 0.35 &  89.0 \scriptsize{±0.8}    & \enskip \textbf{0.8} \scriptsize{±0.4} & 88.0 \scriptsize{±0.4}    & \enskip \textbf{1.5} \scriptsize{±0.2}\\
FisherBCWI     & 0.2 & 81.9 \scriptsize{±0.8}    & \enskip \textbf{1.8} \scriptsize{±0.3}         & 83.5 \scriptsize{±0.2}    & \enskip 2.0 \scriptsize{±0.2} & 0.6 &  85.8 \scriptsize{±0.9}   & \enskip \textbf{0.6} \scriptsize{±0.3} & 85.5 \scriptsize{±0.3}    & \enskip \textbf{0.6} \scriptsize{±0.1}         & 0.2 &  89.2 \scriptsize{±0.8}    & \enskip \textbf{1.0} \scriptsize{±0.4} & 88.1 \scriptsize{±0.4}    & \enskip \textbf{1.9} \scriptsize{±0.3}\\
\midrule
\midrule
SoupBCWI-2     & 0.45 & 81.8 \scriptsize{±0.7}    & \enskip 1.0 \scriptsize{±0.4}         & 83.5 \scriptsize{±0.1}    & \enskip 1.1 \scriptsize{±0.1} & 0.4 &  85.8 \scriptsize{±1.0}   & \enskip 0.6 \scriptsize{±0.3} & 85.6 \scriptsize{±0.4}    & \enskip 0.7 \scriptsize{±0.1}         & 0.45 &  89.1 \scriptsize{±0.6}    & \enskip 0.7 \scriptsize{±0.4} & 88.0 \scriptsize{±0.4}    & \enskip 1.2 \scriptsize{±0.2}\\
SoupBCWI-4     & 0.5 & 81.9 \scriptsize{±0.4}    & \enskip 0.8 \scriptsize{±0.2}         & 83.6 \scriptsize{±0.1}    & \enskip 0.9 \scriptsize{±0.1} & 0.45 &  85.8 \scriptsize{±0.9}   & \enskip 0.5 \scriptsize{±0.2} & 85.4 \scriptsize{±0.4}    & \enskip 0.6 \scriptsize{±0.1}         & 0.45 &  89.1 \scriptsize{±0.8}    & \enskip 0.6 \scriptsize{±0.4} & 88.0 \scriptsize{±0.3}    & \enskip 1.1 \scriptsize{±0.1}\\
SoupBCWI-8     & 0.5 & 81.9 \scriptsize{±0.4}    & \enskip 0.8 \scriptsize{±0.3}         & 83.6 \scriptsize{±0.2}    & \enskip 0.8 \scriptsize{±0.1} & 0.45 &  85.8 \scriptsize{±1.0}   & \enskip 0.5 \scriptsize{±0.2} & 85.4* \scriptsize{±0.3}    & \enskip 0.6 \scriptsize{±0.1}         & 0.45 &  89.1 \scriptsize{±0.7}    & \enskip 0.7 \scriptsize{±0.4} & 88.0 \scriptsize{±0.3}    & \enskip 1.1 \scriptsize{±0.1}\\
SoupBCWI-16     & 0.55 & 81.8 \scriptsize{±0.5}    & \enskip 0.6 \scriptsize{±0.2}         & 83.5 \scriptsize{±0.2}    & \enskip 0.7 \scriptsize{±0.1} & 0.45 &  85.9 \scriptsize{±0.9}   & \enskip 0.5 \scriptsize{±0.2} & 85.4 \scriptsize{±0.4}    & \enskip 0.6 \scriptsize{±0.1}         & 0.5 &  89.1 \scriptsize{±0.8}    & \enskip 0.6 \scriptsize{±0.4} & 87.9 \scriptsize{±0.4}    & \enskip 0.9 \scriptsize{±0.1}\\
\bottomrule
\end{tabular}
}
\caption{Results for the Add\_Data scenario. The trade-off parameter $\lambda$ (or $\alpha$ for BCWI) is tuned on the dev set to be above the accuracy threshold. The threshold is set as 90\% of dev accuracy from old to new model. "*" indicates that the accuracy does not overlap with accuracy of the target model. \textbf{Bold} NFR values have overlapping 95\% confidence intervals with the best value. The old model and SoupBCWI is not under consideration when selecting the best NFR value.}
\label{tab:results_dev_add_data}
\end{table*}
\begin{table*}[t]
\centering
\resizebox{.999\linewidth}{!}{
\begin{tabular}{@{}l|cc@{}cc@{}c|cc@{}cc@{}c|cc@{}cc@{}c@{}}
\toprule
& \multicolumn{5}{c|}{\textbf{MASSIVE}}  & \multicolumn{5}{c|}{\textbf{Banking77}} & \multicolumn{5}{c}{\textbf{AG News}}\\ 
\midrule
 & $\lambda$ & \multicolumn{2}{c}{\textbf{dev}} &  \multicolumn{2}{c|}{\textbf{test}}  & $\lambda$ &  \multicolumn{2}{c}{\textbf{dev}} & \multicolumn{2}{c|}{\textbf{test}}  & $\lambda$ &  \multicolumn{2}{c}{\textbf{dev}} & \multicolumn{2}{c}{\textbf{test}} \\
\midrule
\textbf{Model} &  & \textbf{ACC}\textuparrow &  \textbf{NFR}\textdownarrow  & \textbf{ACC}\textuparrow & \textbf{NFR}\textdownarrow  &  &  \textbf{ACC}\textuparrow & \textbf{NFR}\textdownarrow & \textbf{ACC}\textuparrow &  \textbf{NFR}\textdownarrow  &  &  \textbf{ACC}\textuparrow & \textbf{NFR}\textdownarrow  & \textbf{ACC}\textuparrow & \textbf{NFR}\textdownarrow\\
\midrule
Old Model     & - & 67.1 \scriptsize{±0.5}    & \enskip 0.0 \scriptsize{±0.0}         & 68.8 \scriptsize{±0.1}    & \enskip 0.0 \scriptsize{±0.0} & - &  82.9 \scriptsize{±0.7}    & \enskip 0.0 \scriptsize{±0.0} & 80.0 \scriptsize{±0.3}    & \enskip 0.0 \scriptsize{±0.0}         & - &  67.9 \scriptsize{±0.5}    & \enskip 0.0 \scriptsize{±0.0} & 70.4 \scriptsize{±0.1}    & \enskip 0.0 \scriptsize{±0.0}\\
Target Model     & - & 81.6 \scriptsize{±0.6}    & \enskip 3.9 \scriptsize{±0.5}         & 83.9 \scriptsize{±0.2}    & \enskip 3.2 \scriptsize{±0.2} & - &  89.0 \scriptsize{±0.6}    & \enskip 2.2 \scriptsize{±0.4} & 86.5 \scriptsize{±0.3}    & \enskip 2.8 \scriptsize{±0.2}         & - &  86.8 \scriptsize{±1.5}    & \enskip \textbf{1.5} \scriptsize{±0.7} & 87.9 \scriptsize{±0.5}    & \enskip \textbf{3.5} \scriptsize{±0.6}\\
New Model     & - & 81.5 \scriptsize{±0.6}    & \enskip 3.0 \scriptsize{±0.3}         & 83.8 \scriptsize{±0.2}    & \enskip 2.4 \scriptsize{±0.1} & - &  88.6 \scriptsize{±0.5}   & \enskip 1.7 \scriptsize{±0.6} & 86.3 \scriptsize{±0.3}    & \enskip 1.8 \scriptsize{±0.2}         & - &  87.9 \scriptsize{±0.6}    & \enskip \textbf{1.3} \scriptsize{±0.8} & 87.9 \scriptsize{±0.3}    & \enskip \textbf{4.2} \scriptsize{±0.5}\\
\midrule
ACC Threshold & \multicolumn{2}{c}{\enskip\enskip$\geq$ 80.8} & &&&\multicolumn{2}{c}{\enskip\enskip$\geq$ 88.3} & &&& \multicolumn{2}{c}{\enskip\enskip$\geq$ 86.9} & &&\\
\midrule
\midrule
PriorWD     & 200 & 81.3 \scriptsize{±0.7}    & \enskip 2.1 \scriptsize{±0.3}         & 83.3 \scriptsize{±0.3}    & \enskip 2.1 \scriptsize{±0.2} & 200 &  89.4 \scriptsize{±0.7}   & \enskip \textbf{0.6} \scriptsize{±0.2} & 86.3 \scriptsize{±0.3}    & \enskip 1.1 \scriptsize{±0.1}         & 1e4 &  87.5 \scriptsize{±1.4}    & \enskip \textbf{1.2} \scriptsize{±0.6} & 87.4 \scriptsize{±0.4}    & \enskip 4.3 \scriptsize{±0.4}\\
Mixout     & 0.7 & 81.0 \scriptsize{±0.5}    & \enskip 2.6 \scriptsize{±0.3}         & 83.0 \scriptsize{±0.2}    & \enskip 2.4 \scriptsize{±0.1} & 0.95 &  89.0 \scriptsize{±0.8}   & \enskip \textbf{0.9} \scriptsize{±0.4} & 86.2 \scriptsize{±0.3}    & \enskip 1.2 \scriptsize{±0.1}         & 0.8 &  88.0 \scriptsize{±1.1}    & \enskip \textbf{1.9} \scriptsize{±0.9} & 87.6 \scriptsize{±0.4}    & \enskip 5.0 \scriptsize{±0.5}\\
EWC     & 0.01 & 81.6 \scriptsize{±0.5}    & \enskip 2.2 \scriptsize{±0.3}         & 83.6 \scriptsize{±0.3}    & \enskip 2.0 \scriptsize{±0.1} & 0.01 &  89.3 \scriptsize{±0.7}   & \enskip \textbf{0.6} \scriptsize{±0.2} & 86.4 \scriptsize{±0.3}    & \enskip \textbf{0.9} \scriptsize{±0.1}         & 1e-5 & 88.0 \scriptsize{±0.6} & \enskip \textbf{1.3} \scriptsize{±0.8}          & 87.9 \scriptsize{±0.4}    & \enskip 4.3 \scriptsize{±0.4}\\
\midrule
BCWI     & 0.25 & 81.2 \scriptsize{±0.5}    & \enskip \textbf{1.7} \scriptsize{±0.3}         & 83.2 \scriptsize{±0.2}    & \enskip \textbf{1.4} \scriptsize{±0.1} & 0.35 &  88.8 \scriptsize{±0.5}   & \enskip \textbf{0.8} \scriptsize{±0.3} & 86.0 \scriptsize{±0.4}    & \enskip 1.0 \scriptsize{±0.1}         & 0.1 &  86.9 \scriptsize{±0.6}    & \enskip \textbf{1.1} \scriptsize{±0.6} & 87.6 \scriptsize{±0.3}    & \enskip \textbf{3.6} \scriptsize{±0.4}\\
FisherBCWI     & 0.5 & 81.3 \scriptsize{±0.5}    & \enskip \textbf{1.2} \scriptsize{±0.2}         & 82.9 \scriptsize{±0.2}    & \enskip \textbf{1.2} \scriptsize{±0.1} & 0.7 &  88.5 \scriptsize{±0.5}   & \enskip \textbf{0.6} \scriptsize{±0.2} & 85.7 \scriptsize{±0.5}    & \enskip \textbf{0.7} \scriptsize{±0.1} & 0.05 &  86.9 \scriptsize{±0.7}    & \enskip \textbf{1.0} \scriptsize{±0.6} & 87.5 \scriptsize{±0.2}    & \enskip \textbf{3.3} \scriptsize{±0.4}\\
\midrule
\midrule
SoupBCWI-2     & 0.25 & 81.3 \scriptsize{±0.6}    & \enskip 1.3 \scriptsize{±0.2}         & 83.0 \scriptsize{±0.3}    & \enskip 1.2 \scriptsize{±0.1} & 0.35 &  88.6 \scriptsize{±0.8}   & \enskip 0.8 \scriptsize{±0.4} & 85.8 \scriptsize{±0.4}    & \enskip 0.8 \scriptsize{±0.1}         & 0.05 &  87.3 \scriptsize{±0.8}    & \enskip 1.3 \scriptsize{±0.9} & 87.9 \scriptsize{±0.3}    & \enskip 3.8 \scriptsize{±0.3}\\
SoupBCWI-4     & 0.25 & 81.3 \scriptsize{±0.5}    & \enskip 1.2 \scriptsize{±0.1}         & 82.9 \scriptsize{±0.2}    & \enskip 1.1 \scriptsize{±0.1} & 0.35 &  88.3 \scriptsize{±0.9}   & \enskip 0.8 \scriptsize{±0.4} & 85.8 \scriptsize{±0.3}    & \enskip 0.6 \scriptsize{±0.1}         & 0.05 &  87.4 \scriptsize{±0.9}    & \enskip 1.5 \scriptsize{±0.9} & 87.9 \scriptsize{±0.3}    & \enskip 3.8 \scriptsize{±0.4}\\
SoupBCWI-8     & 0.25 & 81.3 \scriptsize{±0.5}    & \enskip 1.2 \scriptsize{±0.2}         & 82.9 \scriptsize{±0.3}    & \enskip 1.0 \scriptsize{±0.1} & 0.35 &  88.6 \scriptsize{±0.7}   & \enskip 0.7 \scriptsize{±0.3} & 85.8 \scriptsize{±0.3}    & \enskip 0.5 \scriptsize{±0.1}         & 0.1 &  86.9 \scriptsize{±0.8}    & \enskip 1.3 \scriptsize{±0.7} & 87.7 \scriptsize{±0.3}    & \enskip 3.5 \scriptsize{±0.3}\\
SoupBCWI-16     & 0.25 & 81.3 \scriptsize{±0.5}    & \enskip 1.1 \scriptsize{±0.2}         & 82.9 \scriptsize{±0.2}    & \enskip 1.0 \scriptsize{±0.1} & 0.35 &  88.5 \scriptsize{±0.9}   & \enskip 0.5 \scriptsize{±0.3} & 85.8 \scriptsize{±0.3}    & \enskip 0.5 \scriptsize{±0.1}         & 0.05 &  87.7 \scriptsize{±0.8}    & \enskip 1.3 \scriptsize{±0.7} & 87.9 \scriptsize{±0.3}    & \enskip 3.8 \scriptsize{±0.3}\\
\bottomrule
\end{tabular}
}
\caption{Results for the Add\_Classes scenario. The trade-off parameter $\lambda$ (or $\alpha$ for BCWI) is tuned on the dev set to be above the accuracy threshold. The threshold is set as 95\% of dev accuracy from old to new model. "*" indicates that the target accuracy on the test set is not reached. \textbf{Bold} NFR values have overlapping 95\% confidence intervals with the best value. The old model and SoupBCWI is not under consideration when selecting the best NFR value.}
\label{tab:results_dev_add_classes}
\end{table*}

\clearpage
\end{document}